%% file: neurips_2025.tex
\documentclass{article}

\PassOptionsToPackage{authoryear}{natbib}

\usepackage[final]{neurips_2025}

\usepackage[utf8]{inputenc} 
\usepackage[T1]{fontenc}    
\usepackage{hyperref}       
\hypersetup{colorlinks=true, citecolor=blue!50!black, linkcolor=blue!50!black, urlcolor=blue!50!black}
\usepackage{url}            
\usepackage{booktabs}       
\usepackage{amsfonts}       
\usepackage{nicefrac}       
\usepackage{microtype}      
\usepackage{xcolor}         
\usepackage{enumitem}
\setlist{itemsep=1pt, topsep=2pt, parsep=0pt, partopsep=0pt}

\usepackage{amsmath}
\usepackage{amsthm}
\usepackage{tcolorbox}
\usepackage{subcaption}
\usepackage{listings}
\newtcolorbox{examplebox}{
boxrule=0pt,
colback=black!5!white,
sharp corners
}

\title{LLM Layers Immediately Correct Each Other}

\author{%
  Arjun Patrawala \qquad Jiahai Feng \qquad Erik Jones \qquad Jacob Steinhardt \\\\
  University of California, Berkeley\\
  \texttt{\{arjunpatrawala, fjiahai, erjones, jsteinhardt\}@berkeley.edu}
}

\begin{document}

\maketitle

\begin{abstract}

Recent methods in language model interpretability employ techniques such as sparse autoencoders to decompose residual stream contributions into linear, semantically meaningful features. Such methods are commonly interpreted as identifying features that persist in the residual stream and that subsequent layers build upon. We challenge this view by identifying the Transformer Layer Correction Mechanism (TLCM), wherein adjacent transformer layers systematically counteract portions of each other's contributions. TLCM appears in 5 out of 7 major open-source model families and activates across nearly all tokens in diverse texts.
We show that TLCM emerges during pretraining, operates most strongly on contextually dependent tokens, and adaptively calibrates its correction strength based on the preceding layer's output. Using the layer Jacobian, we further show that TLCM selectively corrects specific subspaces while reinforcing others, which we interpret through a ``propose-and-reject'' framework in which layers propose candidate features and subsequent layers selectively remove inappropriate ones. This dynamic suggests that the residual stream at any layer contains transient proposals alongside persistent features, helping explain why SAE feature descriptions often have low specificity, why effective model steering requires extreme feature amplification, and why transcoders hold a theoretical advantage over SAEs.\footnote{Correspondence to \texttt{arjunpatrawala@berkeley.edu}. Code is available at \url{https://github.com/arjunpat/transformer-correction}}
\end{abstract}


\section{Introduction}
\input{sections/0_introduction}

\section{Related Work}
\input{sections/1_related_work}

\begin{figure*}
    \vspace{-30pt}
    \centering
    \includegraphics[width=\linewidth,trim={2mm, 0, 2mm, 0mm},clip]{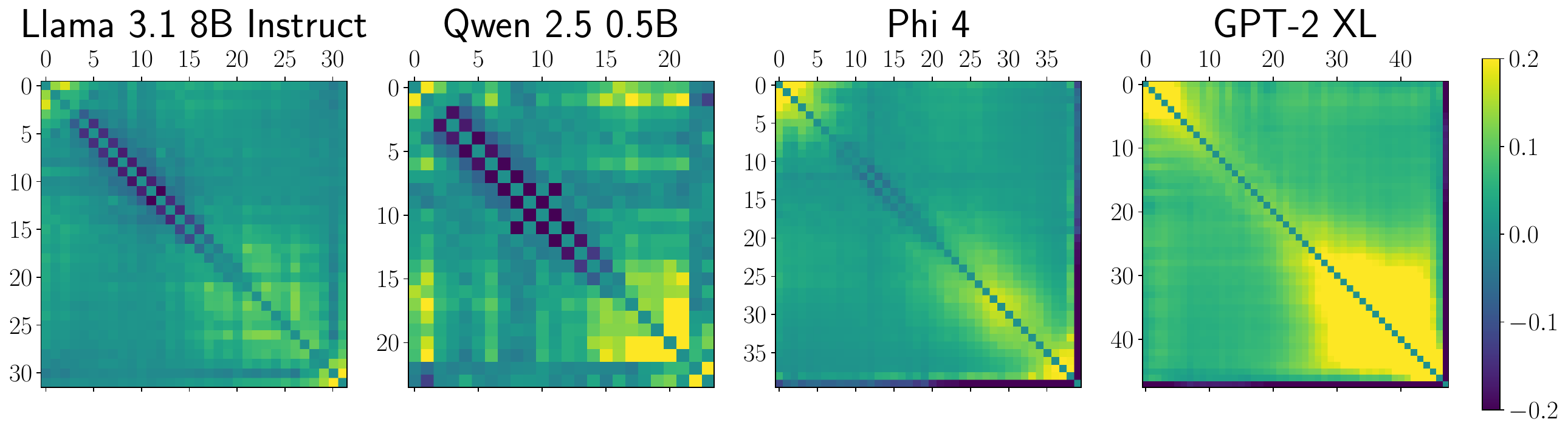}
    \vspace{-12pt}
    \caption{Each panel shows the matrix $\operatorname{clamp}(\mathbf{M}, -0.2, 0.2)$ where $\mathbf{M}[i,j]$ is the average cosine similarity between residual-stream contributions of layers $i$ and $j$ (diagonals zeroed). Negative values (blue) indicate that two layers tend to write opposing vectors; positive values (yellow) indicate reinforcement. In the left two models, adjacent layers exhibit systematic negative similarity, visible as horizontal off-diagonal blue bands in the first two-thirds of layers, revealing the Transformer Layer Correction Mechanism (TLCM). In the right two models, these bands are absent, indicating no consistent adjacent-layer reversals.}
    \label{fig:existence_proof}
\end{figure*}

\section{Background}
\input{sections/2_background}

\section{Transformer Layer Correction Mechanism}
\input{sections/3_tlcm}

\section{TLCM Adaptivity}
\input{sections/4_adaptivity}


\section{Discussion}
\label{sec:discussion}
\input{sections/6_discussion}

\medskip

\bibliography{all.bib}
\bibliographystyle{plainnat}




\newpage
\appendix

\input{sections/appendix}


\newpage
\section*{NeurIPS Paper Checklist}

The checklist is designed to encourage best practices for responsible machine learning research, addressing issues of reproducibility, transparency, research ethics, and societal impact. Do not remove the checklist: {\bf The papers not including the checklist will be desk rejected.} The checklist should follow the references and follow the (optional) supplemental material.  The checklist does NOT count towards the page
limit. 

Please read the checklist guidelines carefully for information on how to answer these questions. For each question in the checklist:
\begin{itemize}
    \item You should answer \answerYes{}, \answerNo{}, or \answerNA{}.
    \item \answerNA{} means either that the question is Not Applicable for that particular paper or the relevant information is Not Available.
    \item Please provide a short (1–2 sentence) justification right after your answer (even for NA). 
\end{itemize}

{\bf The checklist answers are an integral part of your paper submission.} They are visible to the reviewers, area chairs, senior area chairs, and ethics reviewers. You will be asked to also include it (after eventual revisions) with the final version of your paper, and its final version will be published with the paper.

The reviewers of your paper will be asked to use the checklist as one of the factors in their evaluation. While "\answerYes{}" is generally preferable to "\answerNo{}", it is perfectly acceptable to answer "\answerNo{}" provided a proper justification is given (e.g., "error bars are not reported because it would be too computationally expensive" or "we were unable to find the license for the dataset we used"). In general, answering "\answerNo{}" or "\answerNA{}" is not grounds for rejection. While the questions are phrased in a binary way, we acknowledge that the true answer is often more nuanced, so please just use your best judgment and write a justification to elaborate. All supporting evidence can appear either in the main paper or the supplemental material, provided in appendix. If you answer \answerYes{} to a question, in the justification please point to the section(s) where related material for the question can be found.

IMPORTANT, please:
\begin{itemize}
    \item {\bf Delete this instruction block, but keep the section heading ``NeurIPS Paper Checklist"},
    \item  {\bf Keep the checklist subsection headings, questions/answers and guidelines below.}
    \item {\bf Do not modify the questions and only use the provided macros for your answers}.
\end{itemize}


\begin{enumerate}

\item {\bf Claims}
    \item[] Question: Do the main claims made in the abstract and introduction accurately reflect the paper's contributions and scope?
    \item[] Answer: \answerYes{}{} 
    \item[] Justification: We clearly state in the abstract and intro exactly what we were able to show using our experiments. There is a section dedicated to everything we have written in our paper.

\item {\bf Limitations}
    \item[] Question: Does the paper discuss the limitations of the work performed by the authors?
    \item[] Answer: \answerYes{} 
    \item[] Justification: We discuss how we do not see this on a handful of model families and perform some speculation why. In our appendix, we discuss some architectural causes for TLCM (LayerNorm), but also discuss why this might not be true. In our Jacobian experiment, we discuss that directions do not directly correspond to standard features, which limits the strength of results only minorly. In other parts of our work, we caveat that we only see these results strongly from layer $4$ to $20$; or we give the exact dataset or dataset size that we compute across to give a sense for how strong the results are.

\item {\bf Theory assumptions and proofs}
    \item[] Question: For each theoretical result, does the paper provide the full set of assumptions and a complete (and correct) proof?
    \item[] Answer: \answerYes{}{} 
    \item[] Justification: We have one result, and for that reason it is not numbered. Our result and proof is short and is thus presented within the paper. 

    \item {\bf Experimental result reproducibility}
    \item[] Question: Does the paper fully disclose all the information needed to reproduce the main experimental results of the paper to the extent that it affects the main claims and/or conclusions of the paper (regardless of whether the code and data are provided or not)?
    \item[] Answer: \answerYes{}{} 
    \item[] Justification: We generally provide the most important information about reproducibility within the text/content of the paper. However, for some experiments this is intractable, so we include any missing details within the appendix.

\item {\bf Open access to data and code}
    \item[] Question: Does the paper provide open access to the data and code, with sufficient instructions to faithfully reproduce the main experimental results, as described in supplemental material?
    \item[] Answer: \answerNo{}{} 
    \item[] Justification: Our paper mostly consists of short, quick experiments and corresponding matplotlib code that are run on models using HuggingFace transformers. We don't feel it's necessary to release this code as it is reasonably quick to implement once you understand any given experiment, although we are happy to do so if requested.

\item {\bf Experimental setting/details}
    \item[] Question: Does the paper specify all the training and test details (e.g., data splits, hyperparameters, how they were chosen, type of optimizer, etc.) necessary to understand the results?
    \item[] Answer: \answerNA{} 
    \item[] Justification: These details do not apply to our experiments.

\item {\bf Experiment statistical significance}
    \item[] Question: Does the paper report error bars suitably and correctly defined or other appropriate information about the statistical significance of the experiments?
    \item[] Answer: \answerYes{}{} 
    \item[] Justification: For our main results on TLCM, we provide ample information about statistical significance and dedicate a section of our appendix to it. For other experiments, the statistical significance is implied by the high sample size we use, but no error bars or confidence intervals are included.

\item {\bf Experiments compute resources}
    \item[] Question: For each experiment, does the paper provide sufficient information on the computer resources (type of compute workers, memory, time of execution) needed to reproduce the experiments?
    \item[] Answer: \answerNo{} 
    \item[] Justification: We did not include this information because the vast majority of our experiments are not computationally intensive. The most computationally intensive experiment required computing Jacobians of transformer layers, for which we used a single GPU; we detail this in the appendix.
    
\item {\bf Code of ethics}
    \item[] Question: Does the research conducted in the paper conform, in every respect, with the NeurIPS Code of Ethics \url{https://neurips.cc/public/EthicsGuidelines}?
    \item[] Answer: \answerYes{} 
    \item[] Justification: Nothing was harmed in the process of writing this paper. Additionally, we use popular datasets that are publically available.

\item {\bf Broader impacts}
    \item[] Question: Does the paper discuss both potential positive societal impacts and negative societal impacts of the work performed?
    \item[] Answer: \answerNA{} 
    \item[] Justification: We do not believe that there are direct societal impacts of the work performed.
    
\item {\bf Safeguards}
    \item[] Question: Does the paper describe safeguards that have been put in place for responsible release of data or models that have a high risk for misuse (e.g., pretrained language models, image generators, or scraped datasets)?
    \item[] Answer: \answerNA{} 
    \item[] Justification: We do not believe our work poses such risks.

\item {\bf Licenses for existing assets}
    \item[] Question: Are the creators or original owners of assets (e.g., code, data, models), used in the paper, properly credited and are the license and terms of use explicitly mentioned and properly respected?
    \item[] Answer: \answerYes{} 
    \item[] Justification: We primarily use wikitext, which we cite. Additionally, we cite all models discussed or used in the paper.

\item {\bf New assets}
    \item[] Question: Are new assets introduced in the paper well documented and is the documentation provided alongside the assets?
    \item[] Answer: \answerNA{} 
    \item[] Justification: The paper does not release new assets.

\item {\bf Crowdsourcing and research with human subjects}
    \item[] Question: For crowdsourcing experiments and research with human subjects, does the paper include the full text of instructions given to participants and screenshots, if applicable, as well as details about compensation (if any)? 
    \item[] Answer: \answerNA{} 
    \item[] Justification: The paper does not involve crowd-sourcing.

\item {\bf Institutional review board (IRB) approvals or equivalent for research with human subjects}
    \item[] Question: Does the paper describe potential risks incurred by study participants, whether such risks were disclosed to the subjects, and whether Institutional Review Board (IRB) approvals (or an equivalent approval/review based on the requirements of your country or institution) were obtained?
    \item[] Answer: \answerNA{} 
    \item[] Justification: The paper does not involve crowdsourcing nor research with human subjects

\item {\bf Declaration of LLM usage}
    \item[] Question: Does the paper describe the usage of LLMs if it is an important, original, or non-standard component of the core methods in this research? Note that if the LLM is used only for writing, editing, or formatting purposes and does not impact the core methodology, scientific rigorousness, or originality of the research, declaration is not required.
    \item[] Answer: \answerNA{} 
    \item[] Justification: LLMs were only used for help with writing and editing.

\end{enumerate}

\end{document}

%% file: sections/0_introduction.tex
Mechanistic interpretability aims to understand large language models (LLMs) by dissecting the functions 
of their components. This research direction has led to new techniques for building, monitoring, and controlling language models \citep{quirkylm, roger2023benchmarks, li2023inference, feng2024monitoring, pan2024latentqa, xiao2023efficient, wu2024reft}.

A common view in recent interpretability research is that transformer layers additively build upon each other's contributions to enrich representations in the residual stream. For example, analyses of factual recall often characterize successive layers as gradually augmenting entity representations with additional recalled information \citep{geva2023dissectingrecallfactualassociations, rome, nanda2023fact, hernandez2024linearity}. In addition, the view that transformer layers progressively build enriched features motivates recent techniques to extract and interpret such features, including linear probes \citep{alain2016understanding}, the logit lens \citep{nostalgebraist}, sparse autoencoders (SAEs) \citep{bricken2023monosemanticity}, and cross-layer transcoders (CLTs) \citep{ameisen2025circuit, dunefsky2024transcodersinterpretablellmfeature, paulo2025transcodersbeatsparseautoencoders}.


In this work, we demonstrate the Transformer Layer Correction Mechanism (TLCM), in which adjacent transformer layers systematically reverse portions of each other's contributions. Specifically, we find that in 5 of 7 open-weight LLM families (Llama 3, OLMo, Mistral, Gemma, and Qwen2), layer $i+1$ consistently produces a contribution that partially opposes that of the prior layer, layer $i$. TLCM complicates the additive view of the residual stream: adjacent layer contribution vectors can have cosine similarity as low as $-0.4$, indicating that a nontrivial portion of a layer's output is devoted to reversing content written by the preceding layer rather than adding new content. This suggests that features extracted from a single layer's output may not persist in the residual stream---they may be corrected by the subsequent layer, or may themselves serve as corrections of prior content.

To begin, we characterize TLCM through a series of observational experiments in Section \ref{big_sec:tlcm}. First, we show that TLCM is not present at initialization but emerges gradually during pretraining, suggesting that it is learned rather than architectural. Second, we find that TLCM fires most frequently on tokens with high contextual dependency, including numbers, dates, and punctuation, suggesting a role in handling such tokens. Finally, we show that TLCM is the combined effort of both attention and MLP layers.

In Section \ref{big_sec:tlcm_adaptive}, we focus our experiments on TLCM's underlying mechanisms. First, we use causal interventions to show that TLCM is adaptive and directly dependent on the prior layer; in particular layer $i+1$ directly adjusts its correction strength based on the prior layer's contribution strength. Then, we use the transformer layer Jacobian to identify which subspaces are corrected, reinforced, or untouched. Applied to TLCM, we find a subspace corrected 1-for-1 and a smaller subspace reinforced, while non-TLCM interactions show neither pattern.

These findings support a ``propose-and-reject'' interpretation of the forward pass: layers propose candidate features and subsequent layers selectively correct inappropriate ones. Because adjacent layers partially reverse each other's contributions, the residual stream at any given layer contains content that the next layer may partially undo. In Section \ref{sec:discussion}, we discuss how this dynamic connects to three open questions in feature-based interpretability: the low specificity of SAE feature descriptions, the extreme amplification required for model steering, and the advantage of transcoders over SAEs, particularly cross-layer architectures that span adjacent layers.

%% file: sections/1_related_work.tex
\label{sec:recent-work}

\textbf{Feature-based interpretability.}
A common interpretability paradigm interprets the role of a layer in a neural network as the features present in its output.
The linear representation hypothesis posits that neural networks represent concepts as linear features in their activation space \citep{mikolov2013linguistic, olah2020zoom}. 
Feature-based interpretability extracts features using supervised linear probes \citep{alain2016understanding} as well as other more sophisticated techniques that can be unsupervised and/or causal (e.g.~\citet{nostalgebraist, burns2022discovering, geiger2024finding, bricken2023monosemanticity}). 

\textbf{Correction mechanisms.}
Self-repair refers to the phenomenon in which later model components compensate when earlier components are ablated. This has been observed in attention layers \citep{lad2024remarkable, mcgrath2023hydraeffectemergentselfrepair, rushing2024explorations} and in resilience to swapping transformer layers \citep{lad2024remarkable}. Other research has found mechanisms conjectured to help with self-repair, like copy suppression \citep{mcdougall2024copy}. TLCM is more pervasive (occurring multiple times on nearly all tokens) and occurs in an ordinary forward pass, not just under ablation.

\textbf{Residual stream characterization.} The Iterative Inference Hypothesis proposes that transformers progressively refine their latent representations through successive layers \cite{brothers2024uncoveringuncertaintytransformerinference}. Other work builds methods that uncovers semantically-meaningful latents found in the residual stream \cite{belrose2023elicitinglatentpredictionstransformers}. \citet{todd2024function} highlight the process of enrichment: transformer layers construct enriched representations in the residual stream to perform next-token prediction; for example, function vectors encode input-output functions and are placed in the residual stream to induce execution. Additionally, models proactively consolidate entity-related information into the residual stream before it becomes relevant for prediction \cite{hernandez2024linearity, DBLP:conf/acl/LiNA20, hernandez2024inspectingeditingknowledgerepresentations, geva2023dissectingrecallfactualassociations, hendel2023incontextlearningcreatestask, skean2025layerlayeruncoveringhidden}.

%% file: sections/2_background.tex
\textbf{Transformer notation.} Large language models (LLMs) convert a token sequence into a probability distribution over subsequent tokens. Input tokens are first embedded in the $d_\text{m}$-dimensional residual space with positional embeddings, then passed through $n$ transformer layers before being unembedded into token logits.
\begin{align*}
    \mathbf{x}_0 = \operatorname{Embed}(\text{toks}) \qquad
    \mathbf{x}_{i+1} = \operatorname{Layer_i}(\mathbf{x}_i) \qquad
    \text{logits} = \operatorname{Unembed}(\mathbf{x}_n)
\end{align*}
Each transformer layer contains attention and MLP sublayers that produce ``contributions'' to the residual stream:
\begin{align*}
    \mathbf{x}_i' = \mathbf{x}_i + \operatorname{Attn_i}(\mathbf{x}_i) \qquad \qquad
    \mathbf{x}_{i+1} = \mathbf{x}_i' + \operatorname{MLP_i}(\mathbf{x}_i')
\end{align*}
where $\mathbf{x}_i'$ is an intermediate state. The marginal contribution of layer $i$ is defined as $\mathbf{c}_i = \operatorname{Attn_i}(\mathbf{x}_i) + \operatorname{MLP_i}(\mathbf{x}_i + \operatorname{Attn_i}(\mathbf{x}_i))$, making $\mathbf{x}_{i+1} = \mathbf{x}_i + \mathbf{c}_i \label{eq:contribution}$. We define $\mathbf{c}_{i,t}$ to be the contribution of the $i$th layer on token $t$.

\textbf{Feature-based interpretability.} Recent interpretability methods decompose residual stream contributions using sparse autoencoders (SAEs) to extract meaningful features for a residual stream contribution $\mathbf{c}$ \cite{cunningham2023sparseautoencodershighlyinterpretable, gao2024scalingevaluatingsparseautoencoders, templeton2024scaling}:
$$\mathbf{c} = \mathbf{e} + \sum_{i=1}^k \beta_i \mathbf{v}_i$$
where $\mathbf{e}$ is the error, $k$ is a small integer (usually less than 500), and $\mathbf{v}_i$ are the orthogonal feature vectors with activations $\beta_i$. These features can be labeled with semantic meanings and manipulated at inference time to steer model behavior. \cite{marks2025sparsefeaturecircuitsdiscovering, templeton2024scaling, chalnev2024improvingsteeringvectorstargeting}.

\textbf{Models.} We study TLCM across 7 open-weight model families: Llama 3 \cite{grattafiori2024llama3herdmodels}, OLMo \cite{groeneveld2024olmoacceleratingsciencelanguage}, Mistral \cite{jiang2023mistral7b, jiang2024mixtralexperts}, Gemma \cite{gemmateam2024gemmaopenmodelsbased, gemmateam2024gemma2improvingopen}, Qwen2 \cite{yang2024qwen2technicalreport}, GPT-2 \cite{radford2019language}, and Phi \cite{abdin2024phi3technicalreporthighly, abdin2024phi4technicalreport}. Unless otherwise stated, we primarily report results on Llama 3.1 8B ($d_m = 4096$).

%% file: sections/3_tlcm.tex
\label{big_sec:tlcm}




In this section, we introduce the Transformer Layer Correction Mechanism and conduct observational studies to characterize its functioning. We first define it (Section \ref{sec:tlcm-exists}), show that it develops during training (Section \ref{sec:tlcm-develops}), demonstrate how it varies across token types (Section \ref{sec:tlcm-tokens}), and finally demonstrate it occurs via collaboration of both attention and MLP sublayers (Section \ref{sec:tlcm-mlp-attn}).

\subsection{Uncovering the Traces}
\label{sec:tlcm-exists}

If layers progressively enrich the residual stream, a view suggested by prior work, then layer contributions should primarily contribute new content or reinforce existing information. Mechanistically, this might be characterized by the vector layer contributions having roughly zero or positive cosine similarity, corresponding to disjoint contributions and reinforcement, respectively. Negative cosine similarity would indicate something different: one layer \textit{reverses} or \textit{corrects} the other's contribution. Thus, we might expect that over a corpus of in-distribution text, cosine similarity between layer pairs should be predominantly non-negative. Recent work on ViTs supports this, finding positive cosine similarity between layer contributions \cite{jiang2025tracingrepresentationprogressionanalyzing}.

\textbf{Setup.} We test this hypothesis across several open-weight language models. For a given token $t$, we define the similarity matrix $\mathbf{M}_t$, where $\mathbf{c}_{i,t}$ represents the contribution of the $i$-th transformer layer (as defined in equation \ref{eq:contribution}):

$$\mathbf{M}_t[i,j] := \operatorname{cossim}(\mathbf{c}_{i,t}, \mathbf{c}_{j,t}) \text{ for } i \neq j$$
We then average these matrices element-wise across approximately 100,000 tokens, across random documents in WikiText \cite{merity2016pointer} using HuggingFace Transformers \cite{wolf-etal-2020-transformers}: $\mathbf{M} = \frac{1}{n} \sum_t \mathbf{M}_t$. These documents include a variety of languages and code.

\textbf{Results.} Plotted in Figure \ref{fig:existence_proof}, $\mathbf{M}$ reveals a reversing effect, captured by the following observations:
\begin{itemize}
    \item Across this large corpus, adjacent layers (layer $i$ and $i+1$) on average have opposing contributions, evidenced by their negative cosine similarity which averages $\approx -0.2$.
    \item Non-adjacent layers (layer $i$ and $i+j$, $j>1$) \textbf{do not} on average have opposing contributions; their contributions are predominately orthogonal or positively correlated, consistent with the hypothesis above.
\end{itemize}

We term this phenomenon---the systematic partial reversal of layer $i$ by layer $i+1$---the Transformer Layer Correction Mechanism (TLCM). We observe TLCM in 5 of the 7 model families we study, with adjacent-layer cosine similarities typically ranging from $-0.3$ to $-0.15$ and as low as $-0.4$ (Figure \ref{fig:existence_proof}, Appendix \ref{appendix:extended-details-on-tlcm-existence}). TLCM persists across different text types, model scales, and model categories (instruction-tuned and conversational). The correction mechanism is most pronounced in the first two-thirds of each model's layers. TLCM is absent in GPT-2 and the Microsoft Phi models, as shown in Figure \ref{fig:existence_proof}. This absence may stem from their architectures; Phi-3, while based on Llama 2, incorporates dropout blocks after MLP and attention sublayers, similar to GPT-2's design.

\begin{figure}[t]
    \centering
    \includegraphics[width=0.8\linewidth]{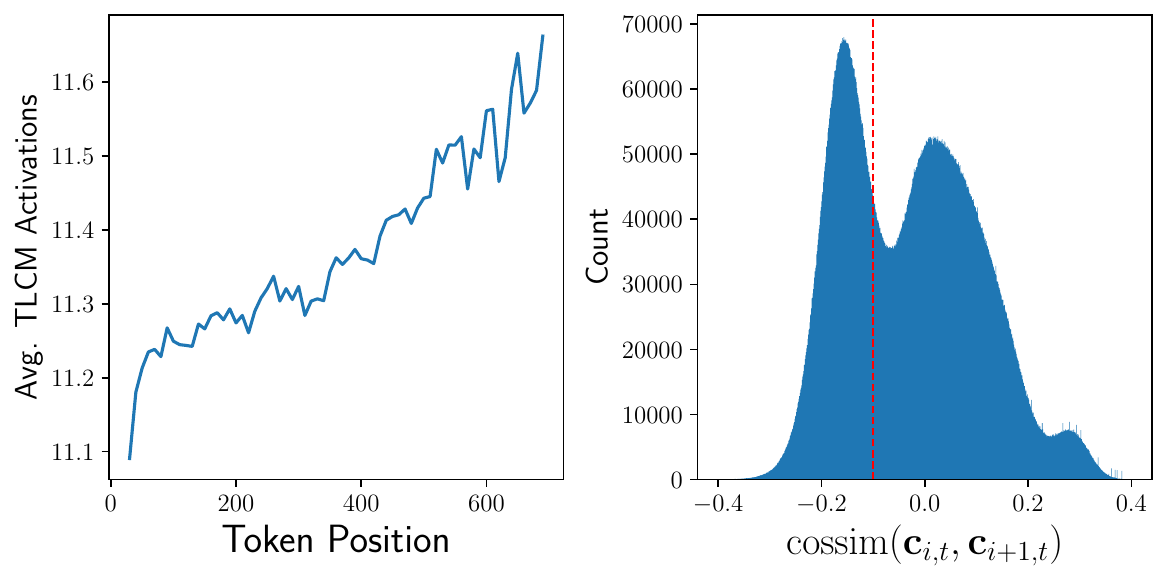}
    \caption{\textbf{Left:} the average number of TLCM activations per token increases with position in the context window. \textbf{Right:} the distribution of adjacent transformer layer similarities is bimodal, with the dotted red line marking the TLCM activation cutoff at $\operatorname{cossim} = -0.1$.}
    \label{fig:avg_tlcm_activations_and_cutoff}
    \vspace{-8pt}
\end{figure}

A natural question is whether the negative cosine similarities we observe form a distinct cluster or are simply the tail of a single distribution. We plot the distribution of $\operatorname{cossim}(\mathbf{c}_{i,t}, \mathbf{c}_{i+1,t})$ across tokens $t$ and layers $i$ in Figure \ref{fig:avg_tlcm_activations_and_cutoff} (right). The distribution is bimodal, with a dominant mode near $-0.2$ (consistent with the average adjacent-layer cosine similarity reported above) and a smaller secondary mode near $+0.1$, separated by a valley around $-0.1$. The presence of a distinct negative mode indicates that these interactions are not rare outliers but a recurring pattern across (token, layer) pairs. Due to the valley between modes at around $-0.1$ cosine similarity, we define any interaction with $-0.1$ or lower cosine similarity to be a TLCM activation.

\textbf{Discussion.} Strong anticorrelation between vectors in a 4096-dimensional space is itself notable. We calibrate the magnitude of these cosine similarities by comparing against two simple null models that assume independence between adjacent layers.

First, suppose both layer outputs are drawn independently from a standard $d$-dimensional normal distribution. Then cosine similarity has expected value 0 and variance $1/d$ (Appendix \ref{appendix:random_normal_iid_vectors}), so at $d = 4096$, a cosine similarity below $-0.1$ corresponds to a 6-sigma event.

Second, suppose both outputs are drawn independently from the empirical distribution of their respective layers, accounting for the possibility that output spaces are low-rank or directionally biased (Appendix \ref{appendix:empirical_mean_and_variance}). Under this null, a typical TLCM instance corresponds to roughly a 3-sigma event.

Both nulls assume independence between adjacent layers, but the architecture explicitly violates this assumption, since layer $i+1$ reads a residual stream that includes layer $i$'s contribution. These comparisons illustrate that the observed cosine similarities are far from what unstructured baselines would predict, but do not establish causality. We provide stronger evidence in Section \ref{sec:tlcm-develops}, where we show TLCM is absent at initialization and emerges during training, and in Section \ref{sec:tlcm-adaptive}, where causal interventions show that layer $i+1$ directly adjusts its output in response to layer $i$.

Beyond the magnitude of the effect, TLCM is puzzling from an efficiency standpoint. Transformer layers may erase information because it is only transiently useful or because deletion frees capacity for new representations. Under either explanation, information written by layer $i$ could be erased by any later layer $j > i$, after multiple downstream layers have had the opportunity to use it. TLCM is more constrained: reversal occurs predominantly in the directly subsequent layer, so written information can only be exploited by a single downstream layer before erasure.

\subsection{TLCM Develops During Training} \label{sec:tlcm-develops}
We next investigate how TLCM emerges over training. In particular, TLCM could be: 
\begin{itemize}
    \item An inherent characteristic of transformer architectures, emerging from their fundamental design rather than through learning.
    \item A transient phenomenon that arises due to, for example, unstable training dynamics.
    \item A mechanism developed during post-training, either by RL or SFT.
\end{itemize}

\textbf{Setup.} To address these possibilities, we study how TLCM develops throughout training. We examine seven checkpoints from each of three fully open-source models: OLMo 7B, OLMo 2 1B, and OLMo 2 7B. For each checkpoint, we compute $\mathbf{M} = \frac{1}{n}\sum_t \mathbf{M}_t$ using a fixed block of text.

\textbf{Results.} Figure \ref{fig:olmo_checkpoints} shows $\mathbf{M}$ computed at four OLMo 7B checkpoints. At step 33,000 (5\% of training), the matrix shows no consistent negative structure on the super-diagonal. By step 135,500 (21\% of training, approximately 0.5 trillion tokens), negative cosine similarities begin to appear between adjacent layers in the middle of the network. This pattern strengthens and extends to more layers through the remainder of training. Appendix Figures \ref{fig:olmo2_1B_checkpoints} and \ref{fig:olmo2_7B_checkpoints} show the same progression in OLMo 2 1B and OLMo 2 7B. These observations indicate that TLCM is
\begin{itemize}
    \item \textit{Learning-induced}: TLCM does not appear in untrained models.
    \item \textit{Stable once learned}: Despite weight decay, TLCM strengthens over training rather than being regularized away.
    \item \textit{Pretraining-derived}: TLCM emerges during pretraining, not SFT or RL.
\end{itemize}


\subsection{TLCM Varies by Token}\label{sec:tlcm-tokens}
We next study how TLCM activations vary across different token classes.

\textbf{Setup.} We curate 100 realistic queries requesting long-form content about science, technology, philosophy, government, history, and other topics. We then generate responses to these queries using Llama 3.1 8B. For each token $t$ in the response, we count the number of corrections for $t$ as: $|\{i \mid \operatorname{cossim}(\mathbf{c}_{i,t}, \mathbf{c}_{i+1,t}) < -0.1\}|$. This corpus contains 79k tokens. To ensure we identify only systematic patterns, we filter out tokens appearing fewer than 20 times and compute the mean number of TLCM activations for each unique token. 

\textbf{Results.} Correction frequencies vary across tokens on Llama 3.1 8B: the lowest-correction tokens average fewer than 8 TLCM activations per token, while the highest average more than 11. Some qualitative examples of low-correction tokens ($<8$ corrections per token on average) include:
\begin{examplebox}
By,  workshops,  indigenous,  vinyl,  implications,  cognitive, -being,  innovative,  innovation,  regulatory,  greenhouse,  mindfulness,  platforms,  trends,  therapy,  example,  learning, iving,  planning,  inclusive,  cloud,  classical,  proposal, -friendly,  sustainability,  biodiversity, ting,  blog,  memo, system, \texttt{<|begin\_of\_text|>}
\end{examplebox}

And some examples of high-correction tokens ($>11$ corrections per token on average):
\begin{examplebox}
202,  Jul, 26, ]$\backslash$n, ,$\backslash$n$\backslash$n, Today, $\backslash$n$\backslash$n,  \$,  Name,  State, [, 201, assistant,  Address, $\backslash$t, 4, $\backslash$n, user,  at, ]$\backslash$n$\backslash$n,  Date,    ,  Title, ], -, Your,  over, Date, \%,  high, ],, )$\backslash$n,  reach,  up,  share, )**,  well,  one,  D,  time, ):,  take,  forward,  from, 1, :,  access,  you, 12,  not,  [, I, City,  need,  low, Thank,  (,  make, 10,  look,  your, 0,  such,  *
\end{examplebox}

We list token-level TLCM activation statistics (mean and standard error of the mean) in Table \ref{table:token_correction_counts_llama_top} and \ref{table:token_correction_counts_llama_bottom}. We also compute results for Gemma 2 2B Instruct across the same 79K tokens, listing full results in Table \ref{table:token_correction_counts_gemma_top} and \ref{table:token_correction_counts_gemma_bottom}.

We find statistically significant differences in the frequency of TLCM activations using two-sided Welch's $t$-tests on per-occurrence TLCM activation counts. For example, on Llama 3.1 8B, the token \texttt{\textbackslash n} averages 12.41 ($\pm 0.10$ SEM, standard error of the mean) TLCM activations, compared with 7.63 ($\pm 0.23$ SEM) for \texttt{sustainability}, a difference of 4.78 activations. Numbers, punctuation, dates, and brackets appear among the tokens with the highest mean activation counts, while English word tokens such as \texttt{community}, \texttt{training}, and \texttt{understanding} have lower mean activation counts. Llama's \texttt{<|begin\_of\_text|>} token and Gemma's \texttt{<bos>} token exhibit the fewest average TLCM activations across the entire corpus.

\textbf{Discussion.} Qualitatively, we observe that low-correction tokens tend to be common English words (e.g., \texttt{sustainability}, \texttt{cognitive}, \texttt{innovation}), while high-correction tokens include numbers, dates, punctuation, and structural tokens (e.g., \texttt{\textbackslash n}, \texttt{[}, \texttt{\$}). One pattern that may account for this split is \textit{contextual dependency}: high-correction tokens tend to be those whose interpretation depends on surrounding context. This includes: 1) Numbers that form larger parts of numbers, where grasping the complete value of the number requires consolidating information \textit{across} tokens 2) Date-related tokens which require surrounding context for complete temporal reference 3) Punctuation marks (e.g., `:') whose semantic role varies with usage context.

Recent work also support the contextual dependency explanation. \citet{lindsey2025biology} showed that newline tokens can be used for planning upcoming tokens, a task that requires contextual processing, while we find that newline tokens and similar (\texttt{\textbackslash n \textbackslash n}, \texttt{:\textbackslash n}, etc.) have high rates of TLCM activation. Intuitively, models may aggregate information into concluding punctuation marks because the causal attention mask prevents them from aggregating information in earlier tokens.

Under the contextual dependency explanation, Llama's \texttt{<|begin\_of\_text|>} (0 activations) and Gemma's \texttt{<bos>} (1 activation) are extreme cases: as the first token in the context window during training, the optimal prediction is simply the unigram distribution from document starts, which obviates contextual processing.

To test this conjecture---that TLCM activations relate to contextual processing---we perform a simple check: we plot the average number of TLCM activations per token at different points in the LLM context window; Intuitively, the 1000th token can attend to 999 preceding tokens while the 5th token can only attend to 4, suggesting we should observe higher TLCM activation rates at later positions. Indeed, across 2000 long WikiText documents, we find that every 1000 tokens of additional context corresponds to approximately 1 additional TLCM activation, as demonstrated in Figure \ref{fig:avg_tlcm_activations_and_cutoff}.

\subsection{Attention and MLPs Alone Do Not Explain TLCM}
\label{sec:tlcm-mlp-attn}

\begin{figure}[t]
    \centering
    \begin{subfigure}[t]{0.57\textwidth}
        \centering
        \includegraphics[width=\textwidth]{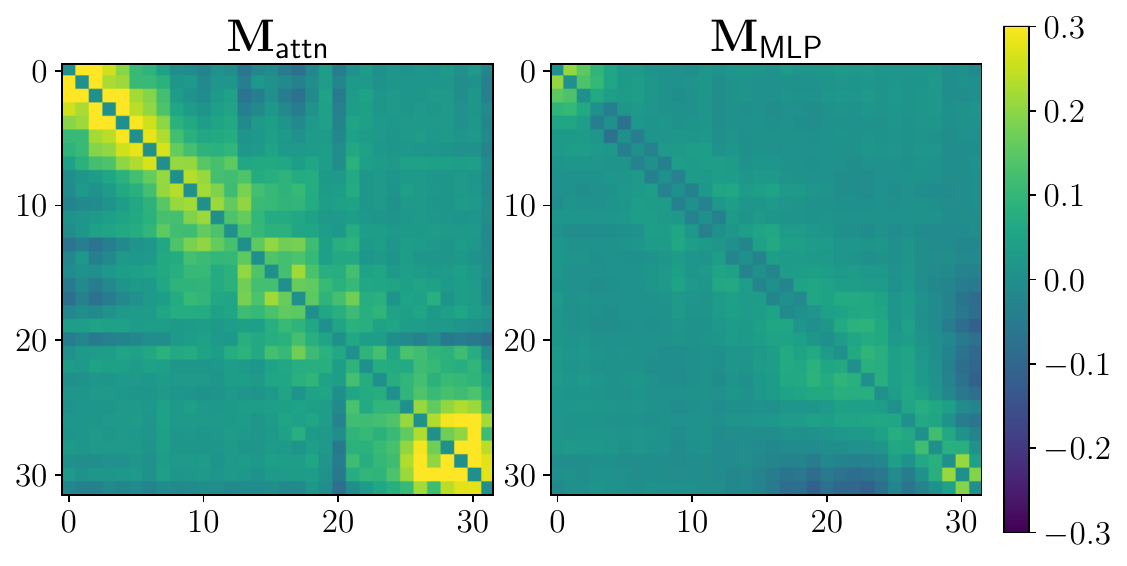}
        \caption{Correlations between attention sublayers, plotted as $\operatorname{clip}(\mathbf{M}_\text{attn}, -0.3, 0.3)$ on the left. Traces of TLCM are seen in the plot of $\operatorname{clip}(\mathbf{M}_\text{MLP}, -0.3, 0.3)$ on the right.}
        \label{fig:m_attn_mlp}
    \end{subfigure}
    \hfill
    \begin{subfigure}[t]{0.4\textwidth}
        \centering
        \includegraphics[width=\textwidth]{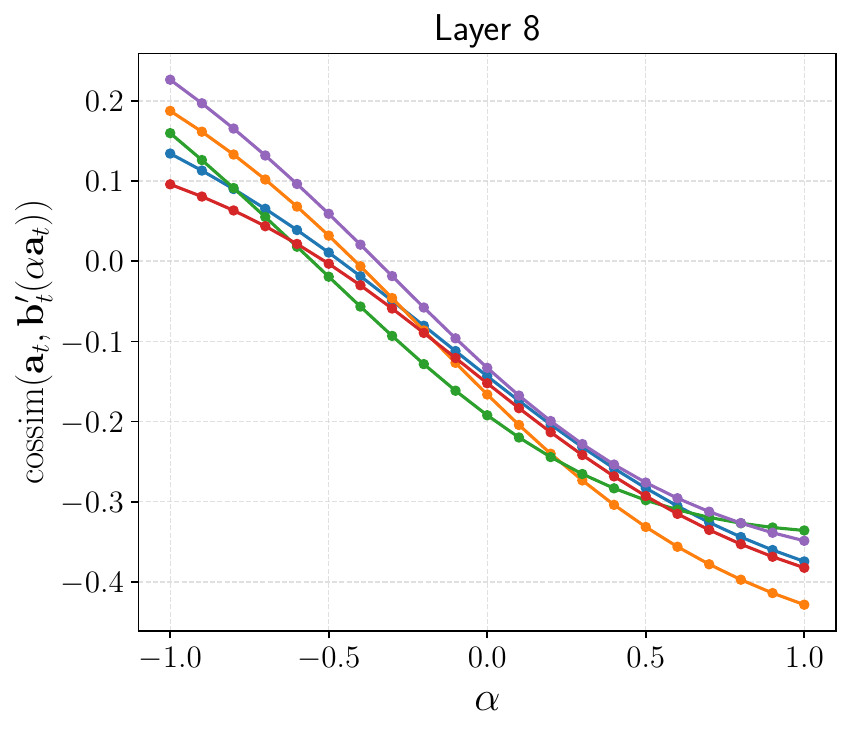}
        \caption{TLCM correction strength varies approximately linearly with the prior layer's contribution, shown for 5 randomly sampled tokens. For negative $\alpha$ values, layers demonstrate compensatory behavior, evidenced by positive cosine similarity.}
        \label{fig:adaptable_correction}
    \end{subfigure}
    \caption{Sublayer-level similarity matrices (a) and TLCM's adaptive correction strength (b).}
    \vspace{-8pt}
\end{figure}

We next investigate the role of MLP and attention sublayers in TLCM. For instance, TLCM could arise purely from MLP-to-MLP interactions, or MLPs could selectively reverse only attention sublayer contributions. To address these questions, we conduct an analysis similar to Section \ref{sec:tlcm-exists}, but at the sublayer level.


\textbf{Setup.} We compute a similarity matrix, instead using the attention contribution $\mathbf{M}_{\text{attn}, t}[i, j] = \operatorname{cossim}(\operatorname{Attn}_i(\mathbf{x}_{i, t}), \operatorname{Attn}_j(\mathbf{x}_{j, t}))$ and plot the average matrix $\mathbf{M}_\text{attn} = \frac{1}{n}\sum_t \mathbf{M}_{\text{attn}, t}$. We also plot the average MLP matrix: $\mathbf{M}_\text{MLP} = \frac{1}{n}\sum_t \mathbf{M}_{\text{MLP}, t}$, with $\mathbf{M}_{\text{MLP}, t}[i, j] = \operatorname{cossim}(\operatorname{MLP}_i(\mathbf{x}_{i, t}'), \operatorname{MLP}_j(\mathbf{x}_{j, t}'))$.

\textbf{Results.} We identify three findings. First, attention sublayers produce positively correlated contributions with other attention sublayers (Figure \ref{fig:m_attn_mlp}). Second, MLP sublayers produce anti-correlated contributions with the subsequent layer's MLP, exhibiting a similar pattern to TLCM. Third, MLP contributions are also anti-correlated with the preceding attention sublayer (Appendix \ref{appendix:extended-details-on-tlcm-existence}). In summary, MLPs produce contributions that are anti-correlated with both the preceding MLP and attention sublayers, suggesting that MLPs are primarily responsible for \textit{executing} corrections.

Since MLPs partially reverse contributions from both the preceding attention and MLP sublayers, the correction mechanism appears to be a transformer layer-level phenomenon. Additionally, our prior experiments suggest contextual dependency plays a role in TLCM's functioning, something only possible if TLCM operates partially through the attention sublayer. For these reasons, we focus our study of TLCM at the transformer layer level.

%% file: sections/4_adaptivity.tex
\label{big_sec:tlcm_adaptive}

In this section, we investigate the mechanisms underlying TLCM. In Section \ref{sec:tlcm-adaptive}, we test whether layer $i+1$'s correction is directly caused by layer $i$'s output, using causal interventions. In Section \ref{sec:jacobian-selective}, we ask whether the correction targets specific subspaces or attenuates the prior layer uniformly, using the transformer layer Jacobian. Finally, we introduce the propose-and-reject hypothesis as a framework that organizes these findings.


\textbf{Notation.} For a particular token $t$, consider the marginal contribution of layer $i$ and layer $i+1$ to be $\mathbf{a}_t$ and $\mathbf{b}_t$, respectively. Recall that the correction mechanism is currently described by a negative cosine similarity between $\mathbf{a}_t$ and $\mathbf{b}_t$ across a variety of layers and tokens. Additionally, observe that the $d_m$-dimensional input to layer $i+1$ would be $\mathbf{x}_{i,t} = \mathbf{x}_{i-1,t} + \mathbf{a}_t$. We define a function $\mathbf{b}_t^\prime$ that captures contribution of layer $i+1$ when the previous layer's output is perturbed by $\mathbf{\Delta}$. Specifically:
\begin{align*}
    \mathbf{b}_t'(\mathbf{\Delta}) := \operatorname{Layer}_{i + 1}(\mathbf{x}_{i,t} + \mathbf{\Delta})
\end{align*}
Note that $\mathbf{b}_t'(\mathbf{0}) = \mathbf{b}_t$, i.e., the original layer contribution.

\subsection{TLCM is Adaptive}
\label{sec:tlcm-adaptive}


In Section \ref{sec:tlcm-exists}, we showed that adjacent-layer anticorrelations are far from what independence-based null models would predict, leaving open whether the cause is a common confounder or direct dependence between the layers.
Here, we use a causal intervention to test whether layer $i+1$'s correction depends directly on layer $i$'s output.


\textbf{Setup.} To understand whether layer $i+1$ is adaptively correcting the contribution of layer $i$, we add some perturbation $\mathbf{\Delta} = \alpha \mathbf{a}_t$, for $\alpha \in [-1, 1]$ to the input of layer $i+1$. Formally, the input to layer $i+1$ is intervened to become $$\mathbf{x}_{i-1,t} +  \mathbf{a}_t + \alpha \mathbf{a}_t = \mathbf{x}_{i-1,t} + (1 + \alpha) \mathbf{a}_t$$

If layer $i+1$ is attuned to correcting $\mathbf{a}_t$, we should expect to see the correction increase as $\alpha$ is scaled up from $0$ and decrease as $\alpha$ is decreased from $0$. To this end, we measure the cosine similarity between $\mathbf{a}_t$ and $\mathbf{b}_t'(\alpha\mathbf{a}_t)$ at $\alpha \in \{-1, -0.9, \ldots, 0.9, 1\}$.

\textbf{Results.}
Increasing $\alpha$ leads to stronger correction. Decreasing $\alpha$ leads to less correction. Figure \ref{fig:adaptable_correction} depicts this near-linear relationship.

When $\alpha = 1$---effectively doubling $\mathbf{a}_t$ in the residual stream---layer $i+1$ responds by adjusting its contribution to further oppose $\mathbf{a}_t$. The relationship between $\alpha$ and correction strength is approximately linear (Figure \ref{fig:adaptable_correction}), and this pattern appears consistently across all layers where TLCM is active, with complete results in Appendix \ref{appendix:tlcm_adaptivity}.

Our experiments also reveal a \textit{compensatory} effect. Specifically, in Figure \ref{fig:adaptable_correction_full} at $\alpha = -1$, the cosine similarity between the adjacent layers becomes \textit{positive}; layer $i+1$ effectively \textit{boosts} the contribution of layer $i$. This effect occurs most strongly in earlier layers. This finding aligns with recent research on self-repair mechanisms that enable models to recover from interventions \citep{mcgrath2023hydraeffectemergentselfrepair}. However, extreme values of $\alpha$ place the model in counterfactual states that do not appear during standard inference or training. Moreover, this compensation effect may stem from self-repair mechanisms unrelated to TLCM.

\textbf{Discussion.} The scaling experiment shows that layer $i+1$'s output responds systematically to perturbations of layer $i$'s contribution, consistent with direct dependence rather than a shared confounder.

One interpretation is that layer $i+1$ identifies layer $i$'s contribution within the full residual stream and adjusts accordingly. However, layer $i+1$'s input is the sum of contributions from all preceding layers, making it unclear how layer $i$'s specific contribution would be isolated. A simpler alternative is that layer $i$'s contribution comprises both components that should persist and components that should be removed, and layer $i+1$ targets the latter. For instance, layer $i$ might write several candidate features, and layer $i+1$---with access to broader context via attention---removes incorrect ones while leaving the rest intact. This motivates two questions:
\begin{enumerate}
    \item Does TLCM uniformly correct the entire previous layer contribution, or does it target specific components?
    \item If TLCM correction is selective, can we identify the most aggressively targeted subspaces?
\end{enumerate}

\subsection{Isolating the Correction Subspace} \label{sec:jacobian-selective}


In this section, we execute an experiment to answer the prior two questions. An open question is \textit{how} the previous layer's contribution is attenuated: is it scaled back uniformly, or are specific subspaces selectively targeted for correction?

The layer Jacobian---the local linearization of both the attention and MLP sublayers---contains information about how the entire layer will respond to input perturbations. The eigenvectors of the symmetrized Jacobian decompose the input space into directions that are corrected, reinforced, or left approximately unchanged by the layer.

\textbf{Notation.} As previously defined, $\mathbf{b}'_t$ is the function representing transformer layer $i+1$. We can linearize this function by computing the Jacobian $\nabla \mathbf{b}'_t(0) = \mathbf{J} \in \mathbb{R}^{d_\text{model} \times d_\text{model}}$ of the transformer layer, such that $\mathbf{b}'_t(\mathbf{\Delta}) \approx \mathbf{b}_t +\mathbf{J}\mathbf{\Delta}$. Finally, let $\overline{\mathbf{J}} := \frac{1}{2}(\mathbf{J} + \mathbf{J}^\top)$ be the symmetrized Jacobian.

\textbf{Claim}. The eigenvectors of $\overline{\mathbf{J}}$ with negative eigenvalues correspond to the corrected subspace, with the eigenvalue dictating the correction strength. In a similar vein, we argue that the positive eigenvalue eigenvectors are reinforced proportional to their eigenvalue. Finally, eigenvectors with near zero eigenvalue are mostly untouched by this layer.

Intuitively, consider a layer input perturbation $\mathbf{\Delta}$ which is an eigenvector of the Jacobian with some negative eigenvalue $\lambda_i < 0$. Observe that the layer's response to this perturbation is approximately $\mathbf{J}\mathbf{\Delta} = \lambda_i \mathbf{\Delta}$ and the net contribution of this perturbation to the residual stream becomes $\mathbf{\Delta} + \lambda_i \mathbf{\Delta}$. For negative $\lambda_i$, this is a correction; if $\lambda_i = -1$, there is zero net contribution from the perturbation, corresponding to a full reversal. Similarly, if the eigenvalue were positive, the subspace would be reinforced. In Appendix \ref{appendix:jacobian_epairs_correct}, we formalize this intuition and show why it's sufficient to consider the eigenpairs of the symmetrized Jacobian $\overline{\mathbf{J}}$, with spectral eigendecomposition $\overline{\mathbf{J}} = \mathbf{Q}\mathbf{V}\mathbf{Q^\top}$.

We can apply this insight to TLCM. Concretely, consider the perturbation from earlier, $\mathbf{a}_t$. We can project $\mathbf{a}_t$ into eigenvector space to obtain $\mathbf{q} = \mathbf{Q}^\top\mathbf{a}_t$. Observe that entry $\mathbf{q}_i$ is the projection onto the $i$-th eigenvector with eigenvalue $\lambda_i$. Suppose we consider the top eigenvectors, namely those with the top fifty $|\mathbf{q}_i|^2$ values; these fifty directions account for $15+$\% of the variance of $\mathbf{a}_t$. Then, we can plot the distribution of the corresponding $\lambda_i$ to understand how TLCM interacts with the top components of $\mathbf{a}_t$.



\begin{figure}[t]
    \vspace{-30pt}
    \centering
    \begin{subfigure}[t]{0.32\textwidth}
        \centering
        \includegraphics[width=\textwidth]{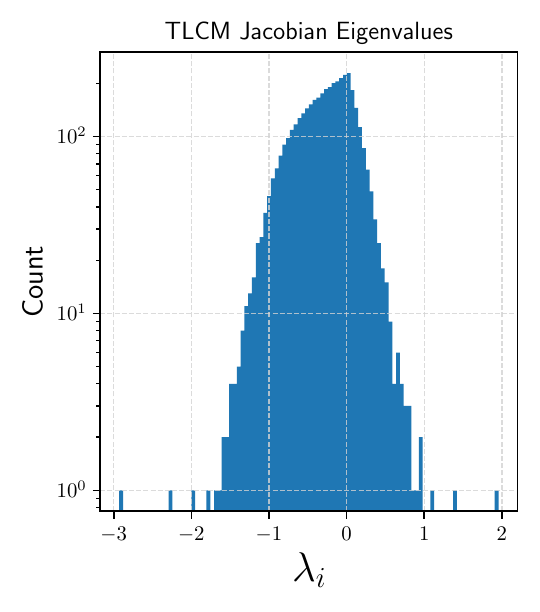}
        \caption{Nearly 3/4 of the 4096 eigenvalues of the Jacobian $\overline{\mathbf{J}}$ are negative.}
        \label{fig:jacobian_evals}
    \end{subfigure}
    \hfill
    \begin{subfigure}[t]{0.65\textwidth}
        \centering
        \includegraphics[width=\textwidth]{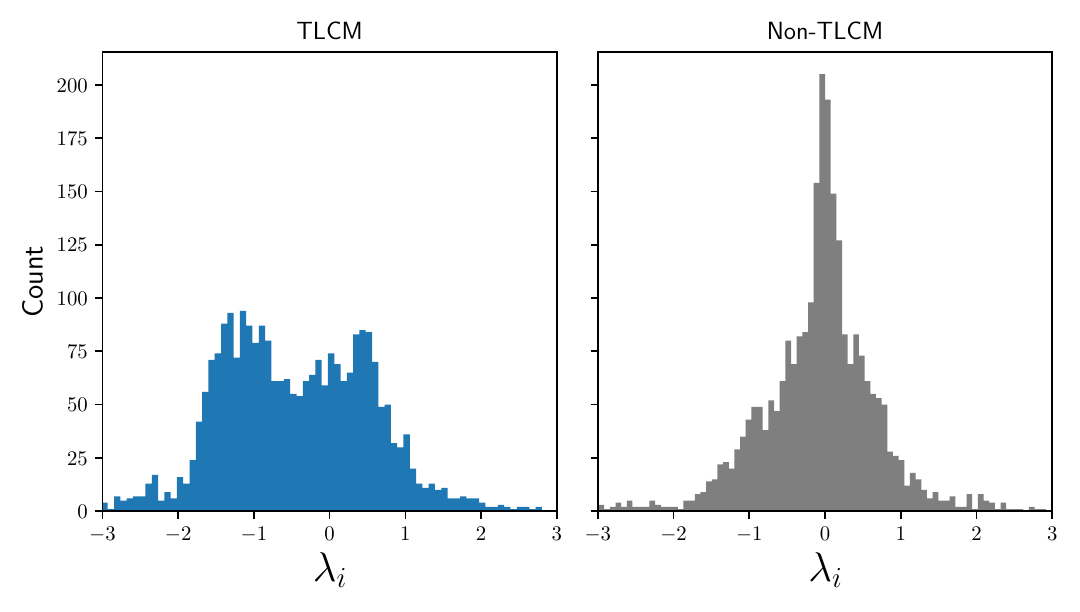}
        \caption{Left: Strongest components of $\mathbf{a}_t$ are corrected or reinforced by the TLCM Jacobian. Right: For non-TLCM Jacobians, $\mathbf{a}_t$'s strongest components are predominantly untouched.}
        \label{fig:jacobian_comparison}
    \end{subfigure}
    \caption{}
    \vspace{-6pt}
\end{figure}

\textbf{Results.} We compute the described plot---the histogram of $\lambda_i$ for top directions of $\mathbf{a}_t$---across 50 Jacobian-$\mathbf{a}_t$ pairs where TLCM is present and 50 pairs where TLCM is not present. Figure \ref{fig:jacobian_comparison} shows the aggregated results. We observe a bimodal eigenvalue distribution for TLCM pairs. The strongest mode hovers near $\lambda = -1$, which means that a unit increase in this direction from the prior layer results in a unit-sized correction.

A mode around $\lambda = -1$ implies correction \textit{rather than} partial attenuation of undesired features; the features are entirely reversed. The other mode hovers around $\lambda = 0.5$, which implies that there are directions being \textit{promoted} by TLCM. This bimodal distribution is inconsistent with uniform attenuation of the previous layer. It targets specific subspaces for correction while leaving others unchanged or reinforcing them. For adjacent layers without TLCM, we observe one strong peak at $\lambda = 0$, suggesting such layers do not interact significantly with the prior layer.

Finally, we observe that about 3000 of the total 4096 eigenvalues of $\overline{\mathbf{J}}$ are negative; see Figure \ref{fig:jacobian_evals}.

\subsection{Propose and Reject Hypothesis}

We have shown that TLCM does not uniformly reverse the prior layer's contribution but rather selectively corrects a subspace of it. To synthesize our findings, we introduce the propose-and-reject hypothesis as a conceptual framework:

\textbf{Propose-and-reject hypothesis (P\&R).} TLCM contributes to feature enrichment through a two-stage process: (1) a layer proposes a set of candidate features, and (2) the subsequent layer, equipped with attention mechanisms to gather context, removes irrelevant features.

P\&R is consistent with our experiments thus far: Attention and MLP layers both play a role in TLCM; TLCM activates more frequently later in the context window; TLCM activates predominately in the first two-thirds of models, which is connected with enrichment; TLCM corrects \textit{only} the prior layer; TLCM appears to perform selective correction rather than uniform attenuation, as shown in Sec. \ref{sec:jacobian-selective}. We note that we do not explicitly test P\&R in this work.

P\&R may be relevant to contextual processing. For example, consider the token ``202'' within ``1,808,202''. During the forward pass, layer $i$ at this token might propose a handful of feature vectors: ``hundred'', ``thousand'', ``million''. Layer $i+1$---equipped with an attention sublayer---can evaluate and eliminate the incorrect features. The incorrect features would form the ``undesirable'' subspace which would be corrected by layer $i+1$. The correct feature, ``million'', would reside in the ``desirable'' subspace and hence remain untouched.






%% file: sections/6_discussion.tex
TLCM connects to several open questions in feature-based interpretability. Under the propose-and-reject hypothesis, the residual stream at any layer $i$ contains both persistent features (those that have survived correction from prior layers) and uncorrected proposals from layer $i$ that layer $i+1$ may reverse. Methods that decompose the residual stream at a single layer cannot distinguish between these two categories. We highlight three connections below.

\textbf{Feature descriptions lack high specificity.} SAEs trained on a single layer's output cannot separate persistent features from soon-to-be-reversed proposals, predicting \textit{misfires}: features that activate on content the model is about to correct. Consistent with this, recent work \cite{templeton2024scaling} observed that over 50\% of activated features from an SAE are labeled by a large LLM as ``Irrelevant'' or ``Only vaguely related'' to the text on which they fire.

\textbf{Effective model steering may need to overcome TLCM's correction.} If TLCM corrections oppose amplified features, then steering interventions may need to exceed the correction capacity to take effect. As shown in Appendix \ref{appendix:tlcm_adaptivity}, TLCM's correction capacity begins to diminish when the prior layer is amplified beyond $2\times$. Recent work found that effective steering requires extreme amplification \textbf{}levels, up to $10\times$ a feature's maximum observed value \cite{templeton2024scaling}, which may reflect the need to overcome correction mechanisms. We propose that intervening on feature directions that TLCM does not target, identified using the Jacobian, could be a promising direction for future work.

\textbf{Transcoders outperform SAEs.} Recent work finds that per-layer transcoders achieve both more faithful reconstruction and more interpretable features than SAEs \cite{paulo2025transcodersbeatsparseautoencoders}. TLCM offers one interpretation of this gap. Unlike SAEs, transcoders model a layer's computation as a function of its input residual stream, so they can in principle represent corrections as input-conditional mechanisms that distinguish features being corrected from features being newly contributed. The cross-layer architecture \cite{ameisen2025circuit} additionally captures dynamics across adjacent layers: a CLT feature that encodes at layer $i$ and decodes to multiple downstream layers can represent both the proposal and its subsequent correction.

Whether TLCM is the most efficient mechanism available, or a consequence of architectural constraints, is an open question. Architectural modifications that reduce TLCM during training, if they lead to lower loss, would suggest the latter. It is also unknown whether TLCM develops in SSM-based sequence models or transformer variations (gated attention, NoPE, MOE).

While the propose-and-reject hypothesis is consistent with our results, it remains a conceptual framework without formalization. One strategy towards validating it might be developing techniques to decompose the correction into semantically meaningful concepts. By employing recent circuit tracing and automated feature labeling techniques, we might be able to semantically explain the correction and its causes.

As a well-characterized and broadly observed phenomenon, TLCM may serve as a useful constraint for researchers building mechanistic accounts of transformer computation.

\newpage

%% file: sections/appendix.tex
\setcounter{table}{0} 
\renewcommand{\thetable}{A\arabic{table}}
\renewcommand{\thefigure}{A\arabic{figure}}
\setcounter{figure}{0}

\section{LayerNorm Blindness to explain TLCM}
\input{sections/5_layernorm_blindness}

\label{appendix:layernorm_blindness}

\begin{figure}
    \centering
    \includegraphics[width=\linewidth]{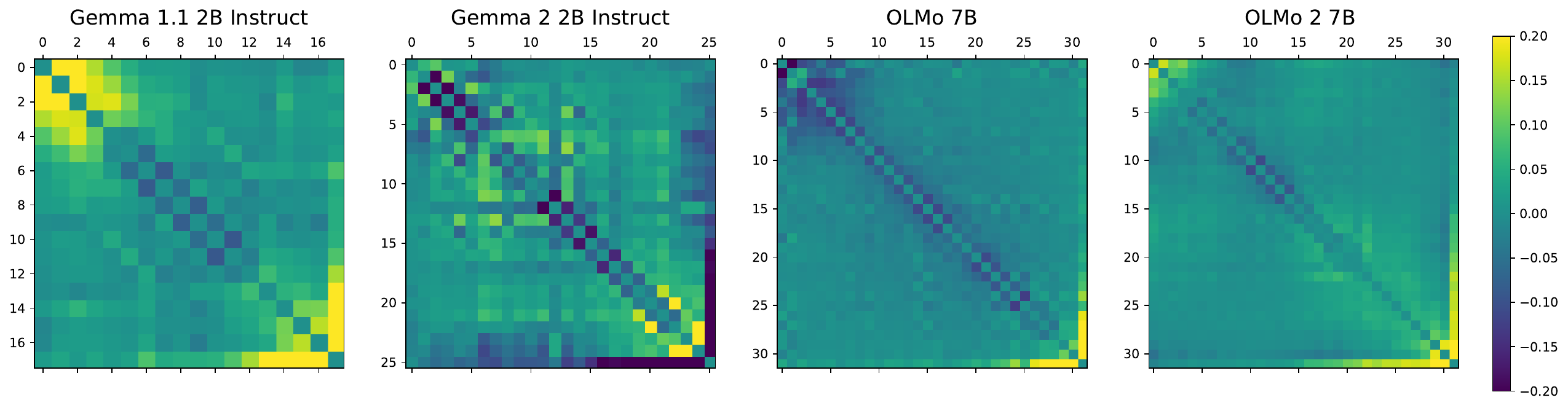}
    \caption{TLCM is consistently found on models with different approaches to LayerNorm. Gemma 2 with post-LayerNorm has more pronounced TLCM. OLMo 2, which replaces pre-LayerNorm entirely exhibits TLCM. For visual clarity, we plot $\operatorname{clamp}(\mathbf{M}, -0.2, 0.2)$ and zero the diagonals.}
    \label{fig:rmsnorm}
\end{figure}

\section{Jacobian Eigendecomposition Locates Corrected Subspaces}
\label{appendix:jacobian_epairs_correct}

We restate notation for convenience.

\textbf{Notation.} Consider $\mathbf{b}'_t$ to be the function representing transformer layer $i+1$. We can linearize this function by computing the Jacobian $\nabla \mathbf{b}'_t(0) = \mathbf{J} \in \mathbb{R}^{d_\text{model} \times d_\text{model}}$ of the transformer layer, such that $\mathbf{b}'_t(\mathbf{\Delta}) \approx \mathbf{b}_t +\mathbf{J}\mathbf{\Delta}$. Finally, let $\overline{\mathbf{J}} := \frac{1}{2}(\mathbf{J} + \mathbf{J}^\top)$ be the symmetrized Jacobian.

\textbf{Claim}. The eigenvectors of $\overline{\mathbf{J}}$ with negative eigenvalues correspond to the corrected subspace, with the eigenvalue dictating the correction strength. In a similar vein, we argue that the positive eigenvalue eigenvectors are reinforced proportional to their eigenvalue. Finally, eigenvectors with near zero eigenvalue are mostly untouched by this layer.

\textbf{Proof.} An input perturbation to layer $i+1$ is ``corrected'' if $\operatorname{cossim}(\mathbf{b}'_t(\mathbf{\Delta}) - \mathbf{b}_t, \mathbf{\Delta}) < 0$. Observe that this condition is true if and only if $\langle \mathbf{\mathbf{b}'_t(\mathbf{\Delta}) - \mathbf{b}_t}, \mathbf{\Delta} \rangle < 0$. For a small enough perturbation, the condition is equivalent to $\langle \mathbf{J\Delta}, \mathbf{\Delta} \rangle < 0$. Because $\langle \mathbf{J\Delta}, \mathbf{\Delta} \rangle = \langle \mathbf{\Delta}, \mathbf{J\Delta} \rangle$, the condition becomes:

\begin{align}
\langle \mathbf{J\Delta}, \mathbf{\Delta} \rangle = \frac{1}{2}\left(\langle \mathbf{J\Delta}, \mathbf{\Delta} \rangle + \langle \mathbf{\Delta}, \mathbf{J\Delta} \rangle\right) = \langle \frac{1}{2}(\mathbf{J} + \mathbf{J}^\top)\mathbf{\Delta}, \mathbf{\Delta} \rangle = \langle \mathbf{\overline{J}\Delta}, \mathbf{\Delta} \rangle < 0 \label{eq:sym_jacobian}
\end{align}

Moreover, by the spectral theorem, $\overline{\mathbf{J}}$ has an eigendecomposition $\overline{\mathbf{J}} = \mathbf{QVQ^\top}$ with unitary eigenvector matrix $\mathbf{Q} \in \mathbb{R}^{d_\text{m} \times d_\text{m}}$ and eigenvalues are $\operatorname{diag}(\mathbf{V})$. Plugging this into inequality \ref{eq:sym_jacobian}:

\begin{align}
\langle \overline{\mathbf{J}} \mathbf{\Delta}, \mathbf{\Delta} \rangle = \langle \mathbf{QVQ^\top} \mathbf{\Delta}, \mathbf{\Delta} \rangle &= \mathbf{\Delta}^\top \mathbf{QVQ^\top} \mathbf{\Delta} = (\mathbf{Q}^\top \mathbf{\Delta})^\top \mathbf{V} (\mathbf{Q}^\top \mathbf{\Delta}) < 0 \label{eq:neg_evals}
\end{align}

In summary, we have reduced our original condition for a corrected perturbation, $\operatorname{cossim}(\mathbf{b}'_t(\mathbf{\Delta}) - \mathbf{b}_t, \mathbf{\Delta}) < 0$, into something more tractable to analyze: $(\mathbf{Q}^\top \mathbf{\Delta})^\top \mathbf{V} (\mathbf{Q}^\top \mathbf{\Delta}) < 0$. Observe that $\mathbf{Q}^\top \mathbf{\Delta} \in \mathbb{R}^{d_\text{model}}$ is a vector of the perturbation projected into the Jacobian's eigenvector space. In, \ref{eq:neg_evals}, each projected component is squared and multiplied by the corresponding eigenvalue. Thus, if a perturbation decomposes heavily onto an eigenvector with a negative eigenvalue, our original condition for "correction" will be satisfied. \qed




\section{Likelihood of Negative Cosine Similarity}

\subsection{Random Normal IID Vectors}
\label{appendix:random_normal_iid_vectors}
Consider two random vectors: $\mathbf{u}, \mathbf{v} \sim \mathcal{N}(\mathbf{0}, \sigma^2 \mathbf{I}_d)$.

\begin{align*}
\mathbb{E}[\operatorname{cossim}(\mathbf{u}, \mathbf{v})] &= \mathbb{E}\left[\frac{\langle \mathbf{u}, \mathbf{v} \rangle}{\|\mathbf{u}\|_2 \|\mathbf{v}\|_2}\right]\\
&= \mathbb{E}\left[ \frac{\sum_i \mathbf{u}_{i} \mathbf{v}_{i}}{\sqrt{\sum_i \mathbf{u}_i^2} \sqrt{\sum_i \mathbf{v}_i^2}} \right]\\
&= \mathbb{E}\left[ \sum_j \frac{\mathbf{u}_j}{\sqrt{\sum_i \mathbf{u}_i^2}} \frac{\mathbf{v}_j}{\sqrt{\sum_i \mathbf{v}_i^2}} \right]\\
&= \sum_j \mathbb{E}\left[ \frac{\mathbf{u}_j}{\sqrt{\sum_i \mathbf{u}_i^2}} \right] \mathbb{E}\left[ \frac{\mathbf{v}_j}{\sqrt{\sum_i \mathbf{v}_i^2}}\right]\\
&= 0
\end{align*}

The last expectations must be 0 because $\frac{\mathbf{u}_j}{\sqrt{\sum_i \mathbf{u}_i^2}}$ is distributed the same as $\frac{-\mathbf{u}_j}{\sqrt{\sum_i \mathbf{u}_i^2}}$. To calculate the variance, first observe the following fact:

\begin{align*}
\mathbb{E}\left[ \sum_{j=1}^d  \frac{\mathbf{u}_j^2}{\sum_i \mathbf{u}_i^2} \right] = 1 \\
\sum_{j=1}^d \mathbb{E}\left[   \frac{\mathbf{u}_j^2}{\sum_i \mathbf{u}_i^2} \right] = 1 \\
d\mathbb{E}\left[   \frac{\mathbf{u}_1^2}{\sum_i \mathbf{u}_i^2} \right] = 1\\ \mathbb{E}\left[   \frac{\mathbf{u}_1^2}{\sum_i \mathbf{u}_i^2} \right] = \frac{1}{d}
\end{align*}

Using this identity, the variance is:

\begin{align*}
\operatorname{Var}[\operatorname{cossim}(\mathbf{u}, \mathbf{v})] &= \mathbb{E}[\operatorname{cossim}(\mathbf{u}, \mathbf{v})^2]\\
&= \mathbb{E}\left[ \frac{(\sum_i \mathbf{u}_{i} \mathbf{v}_{i})(\sum_i \mathbf{u}_{i} \mathbf{v}_{i})}{\sum_i \mathbf{u}_i^2 \sum_i \mathbf{v}_i^2} \right]\\
&= \mathbb{E}\left[ \frac{\sum_{i,j} \mathbf{u}_{i} \mathbf{v}_{i} \mathbf{u}_{j} \mathbf{v}_{j}}{\sum_i \mathbf{u}_i^2 \sum_i \mathbf{v}_i^2} \right]\\
&= \mathbb{E}\left[ \frac{\sum_{i} \mathbf{u}_{i}^2 \mathbf{v}_{i}^2}{\sum_i \mathbf{u}_i^2 \sum_i \mathbf{v}_i^2} \right]\\
&= \sum_{j=1}^d \mathbb{E}\left[ \frac{\mathbf{u}_j^2}{\sum_i \mathbf{u}_i^2} \right] \mathbb{E}\left[ \frac{\mathbf{v}_j^2}{\sum_i \mathbf{v}_i^2} \right]\\
&= \sum_{j=1}^d \frac{1}{d} \cdot \frac{1}{d} = \frac{1}{d}\\
\end{align*}

\subsection{Experimental Mean and Standard Deviation}
\label{appendix:empirical_mean_and_variance}


We find that the contributions of a particular layer are anisotropic; they cluster around approximately 500-800 of the 4096 dimensions of Llama's residual stream.

\begin{figure}
    \centering
    \includegraphics[width=0.9\textwidth]{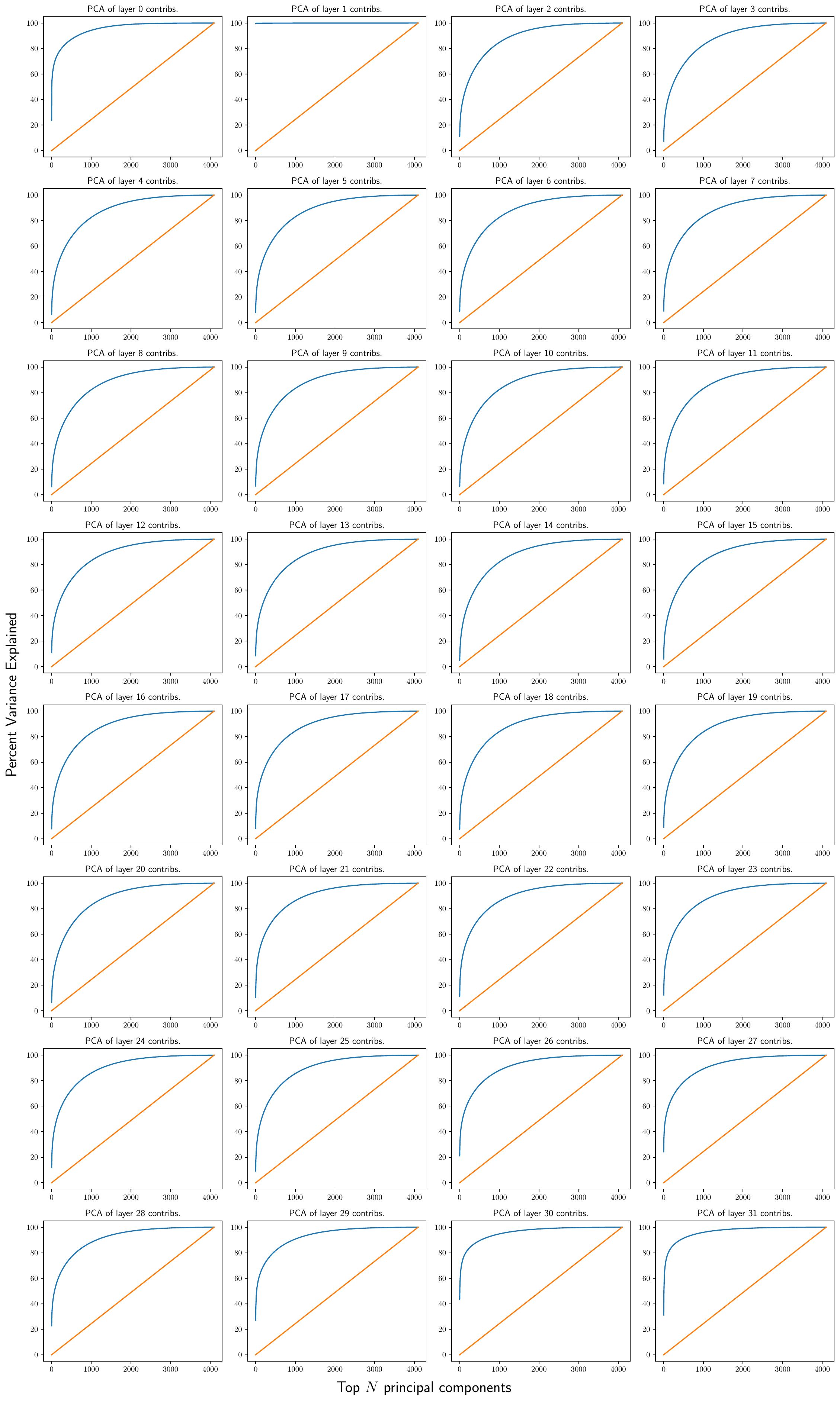}
    \caption{Percent variance explained by the top $n$ principal components vs $n$, graphed on the blue line. The orange line is what we would expect if contributions were isotropic. For layers where TLCM is most active (4 to 20), the first 500-800 principal components explain 80\% of the variance in contributions, demonstrating that contributions are anisotropic. }
    \label{fig:contributions_pca}
\end{figure}

Specifically, we isolate the contribution vector for layer $i$ across 4096 tokens from wikitext: $\{\mathbf{c}_{i, t_1}, \mathbf{c}_{i, t_2}, \ldots, \mathbf{c}_{i, t_{4096}} \}$. After computing the SVD of each set, we plot percent variance explained by top $n$ principal components vs. $n$ in Figure \ref{fig:contributions_pca}.

Due to this anisotropy, we calculate the statistical significance of the TLCM cutoff ($-0.1$ cosine similarity) empirically. We compute the mean and variance of cosine similarity across contributions of adjacent layers on \textit{different} tokens. More formally, we compute the mean and variance of $\operatorname{cossim}(\mathbf{c}_{i,t_1}, \mathbf{c}_{i+1, t_2})$ for $t_1 \neq t_2$ and $4 \leq i < 20$. We find it has has mean $-0.00375$ and standard deviation is $0.03267$, meaning that our cutoff of $-0.1$ is conservatively a $3\sigma$ event; most TLCM events occur at lower cosine similarities ($-0.15$ to $-0.25$).

\begin{figure}
    \centering
    \includegraphics[width=\linewidth]{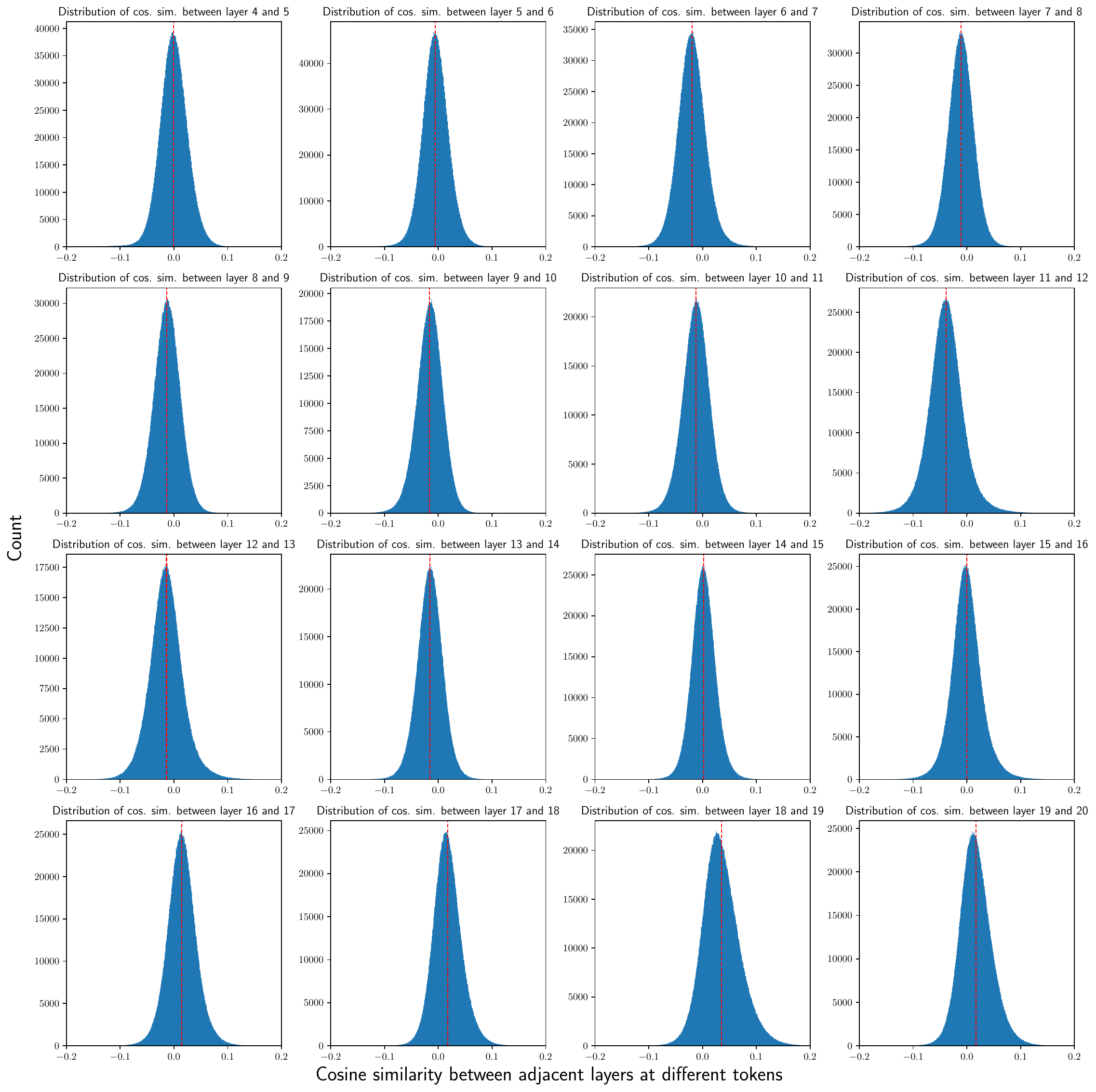}
    \caption{The empirical distribution of $\operatorname{cossim}(\mathbf{c}_{i,t_1}, \mathbf{c}_{i+1, t_2}), t_1 \neq t_2$ across 2000 tokens for layers $4 \leq i < 20$, corresponding to approximately 2 million samples per histogram. The red dotted line corresponds to the mean.}
    \label{fig:inter_layer_cossim}
\end{figure}

We plot these distributions by layer in Figure \ref{fig:inter_layer_cossim}.

\section{Details on Jacobian Experiments}
\label{appendix:jacobians_details}

\subsubsection{Jacobian Sanity Check}

We have approximated our response function $\mathbf{b}_t'$ using the Taylor expansion.

\begin{align*}
\mathbf{b}_t'(\mathbf{\Delta}) &= \mathbf{b}_t'(0) + \mathbf{J}\mathbf{\Delta} + \mathcal{O}(\mathbf{\Delta}^2)
\end{align*}

Here we aim to confirm that the Jacobian is appropriately representative of the transformer layer within a reasonable regime. Observe that the error of this approximation is $\mathbf{b}_t'(\mathbf{\Delta}) - \mathbf{b}_t'(0) - \mathbf{J}\mathbf{\Delta}$, and we thus denote the percent error of the approximation as follows:

$$\operatorname{err}(\mathbf{\Delta}) = \frac{\|\mathbf{b}_t'(\mathbf{\Delta}) - \mathbf{b}_t'(0) - \mathbf{J}\mathbf{\Delta}\|}{\|\mathbf{b}_t'(\mathbf{\Delta}) - \mathbf{b}_t'(0)\|}$$

\begin{figure}
    \centering
    \includegraphics[width=0.9\textwidth]{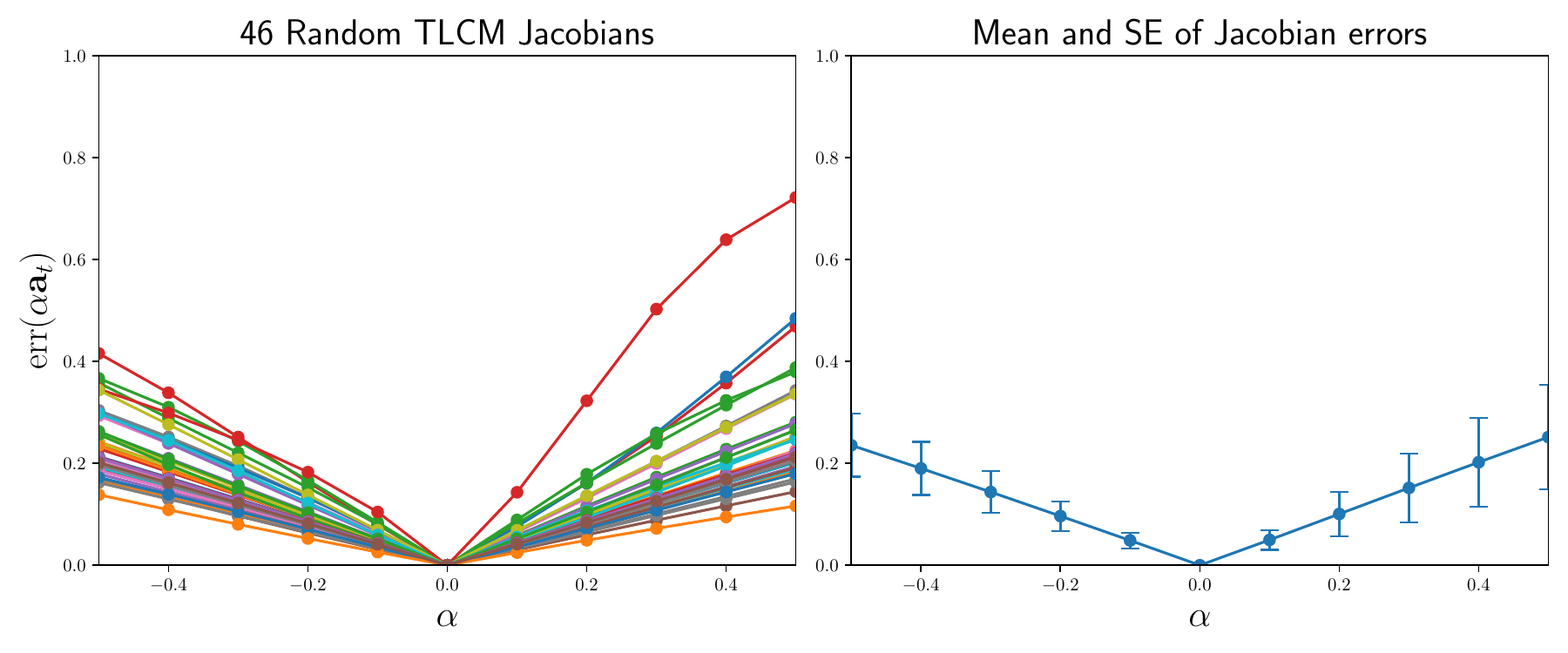}
    \caption{The transformer layer Jacobian approximation has consistently around 5\% error for perturbations ($|\alpha| < 0.1$) to the following layer, making it useful for decomposing directions as described in Sec. \ref{sec:jacobian-selective}.}
    \label{fig:jacobian_sanity_check}
\end{figure}

We plot the percent error error of this approximation across 46 randomly selected TLCM Jacobians for different values of $\mathbf{\Delta} = \alpha \mathbf{a}_t$, $\alpha \in \{-0.5, -0.4, -0.3, \ldots, 0.3, 0.4, 0.5\}$. Shown in Figure \ref{fig:jacobian_sanity_check}, we find that the Jacobian is a good approximation within this regime. For $|\alpha| < 0.1$, there is consistently around 5\% error, which we believe is sufficient for our Jacobian-based analysis in Sec. \ref{sec:jacobian-selective}.

\newpage

\section{Extended Details on TLCM Existence}
\label{appendix:extended-details-on-tlcm-existence}

\begin{figure}
    \centering
    \includegraphics[width=\textwidth]{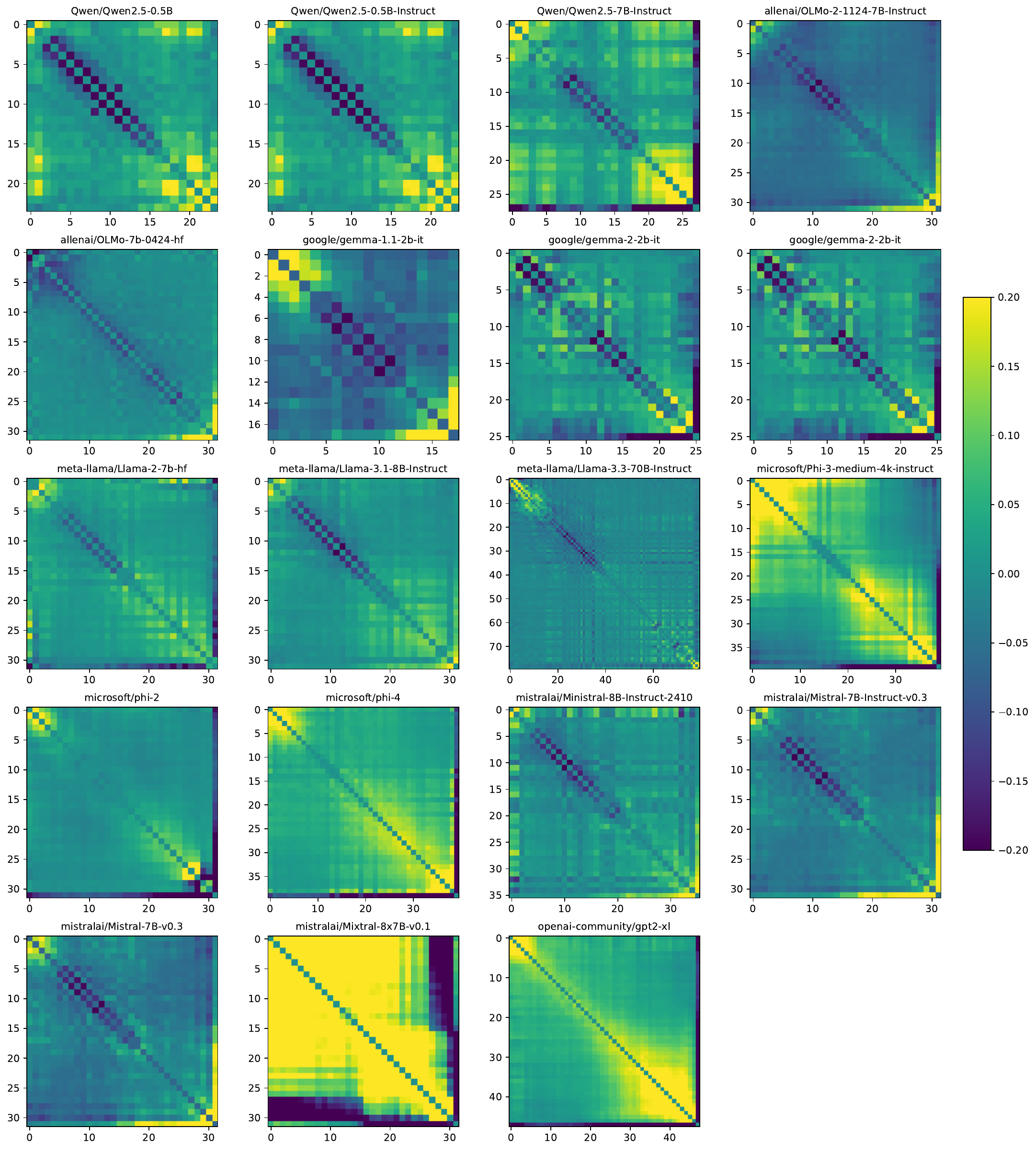}
    \caption{We plot $\operatorname{clamp}(\mathbf{M}, -0.2, 0.2)$ for four models, zeroing diagonal entries. We computed $\mathbf{M}$ across a variety of models, pulled from Huggingface Transformers.}
    \label{fig:matrix_figures}
\end{figure}

In Figure \ref{fig:matrix_figures}, we plot TLCM's existence across many HuggingFace models using the same technique as described in Sec. \ref{sec:tlcm-exists} of the main body.

We previously demonstrated that MLPs exhibit anti-correlations with each other, while attentions do not. We additionally find that MLPs correct attentions (both within and between transformer layers) and that attentions correct MLPs from prior layers. This could be due to a common low-dimensional subspace used by both units for communication. Thus, MLPs correct both prior attention and MLPs; attentions correct just prior MLPs. Altogether, this suggests that MLPs are more responsible for TLCM's correction, but both units are involved.

\begin{figure}
    \centering
    \includegraphics[width=0.6\linewidth]{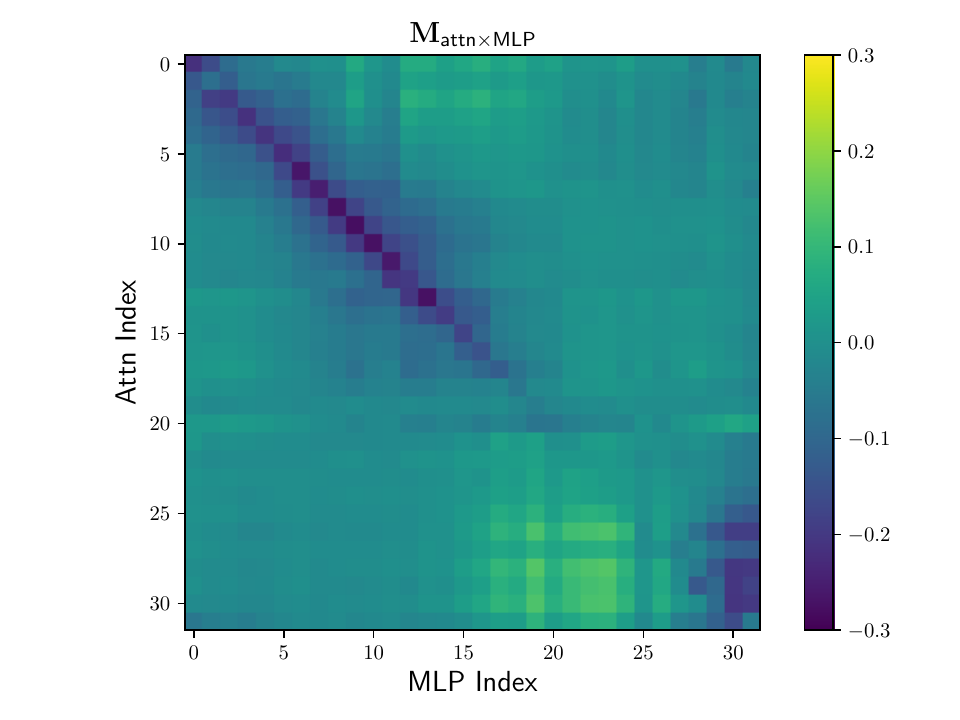}
    \caption{MLPs correct prior attentions, and attentions correct prior MLPs.}
    \label{fig:m_attn_cross_mlp}
\end{figure}

Specifically, we plot $\mathbf{M}_{\text{attn}\times\text{MLP}} = \frac{1}{n}\sum_t \mathbf{M}_{\text{attn}\times\text{MLP},t}$ where $$\mathbf{M}_{\text{attn}\times\text{MLP},t}[i,j] := \operatorname{cossim}(\operatorname{Attn}_i(\mathbf{x}_{i, t}), \operatorname{MLP}_j(\mathbf{x}_{j,t}))$$ See Figure \ref{fig:m_attn_cross_mlp}.

\section{Extended Details on TLCM Adaptivity}
\label{appendix:tlcm_adaptivity}

\subsection{TLCM is Adaptive Across Layers}

\begin{figure}
    \centering
    \includegraphics[width=\linewidth]{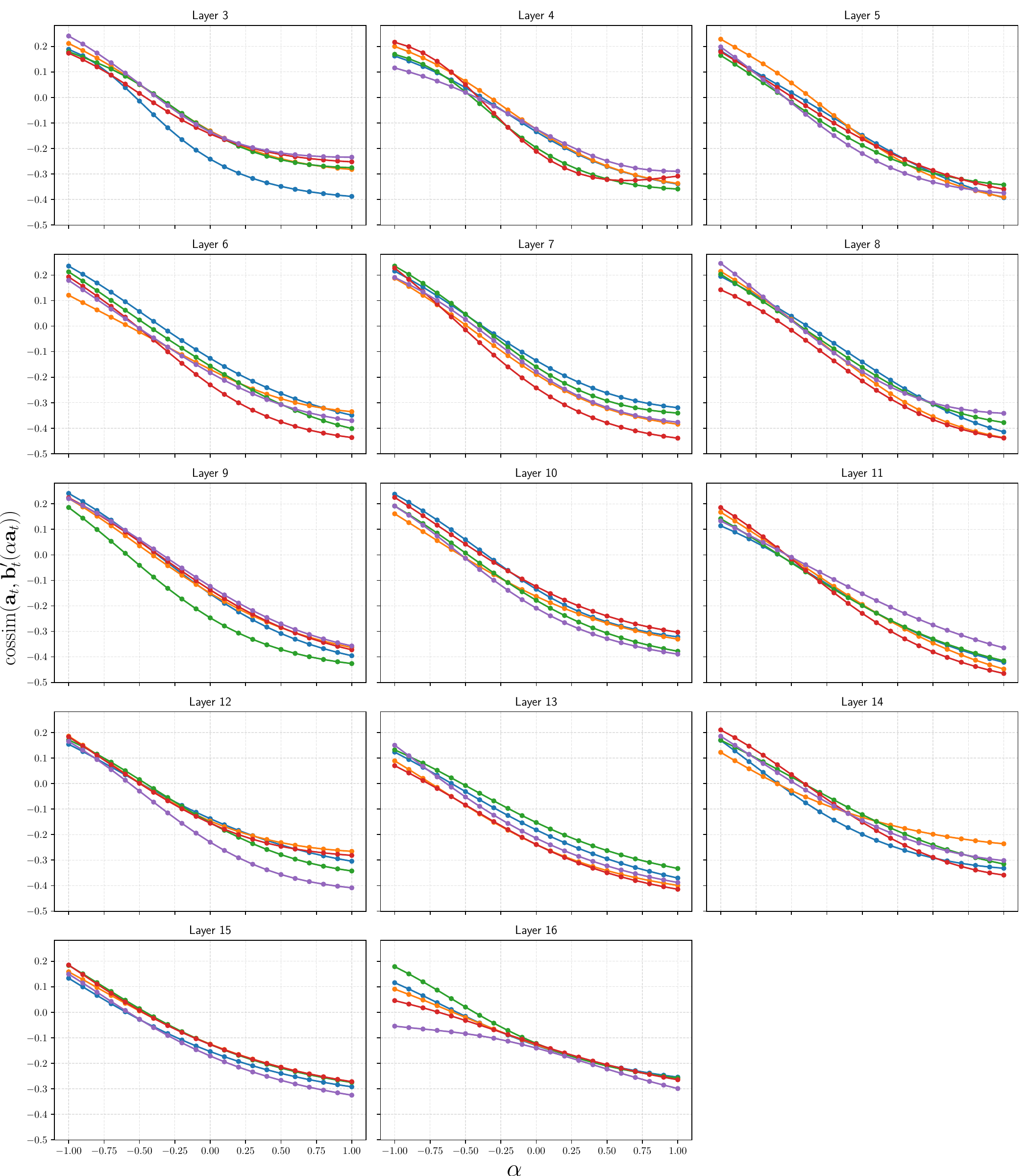}
    \caption{TLCM correction strength varies approximately linearly with the prior layer's contribution. We plot 5 random corrections across a variety of layers in Llama 3.1 8B.}
    \label{fig:adaptable_correction_full}
\end{figure}

In Figure \ref{fig:adaptable_correction_full}, we plot TLCM's adaptive correction across different layers using the same technique as described in Sec. \ref{sec:tlcm-exists}.

\subsection{TLCM Correction Capacity Diminishes at High $\alpha$}

\begin{figure}
    \centering
    \includegraphics[width=\linewidth]{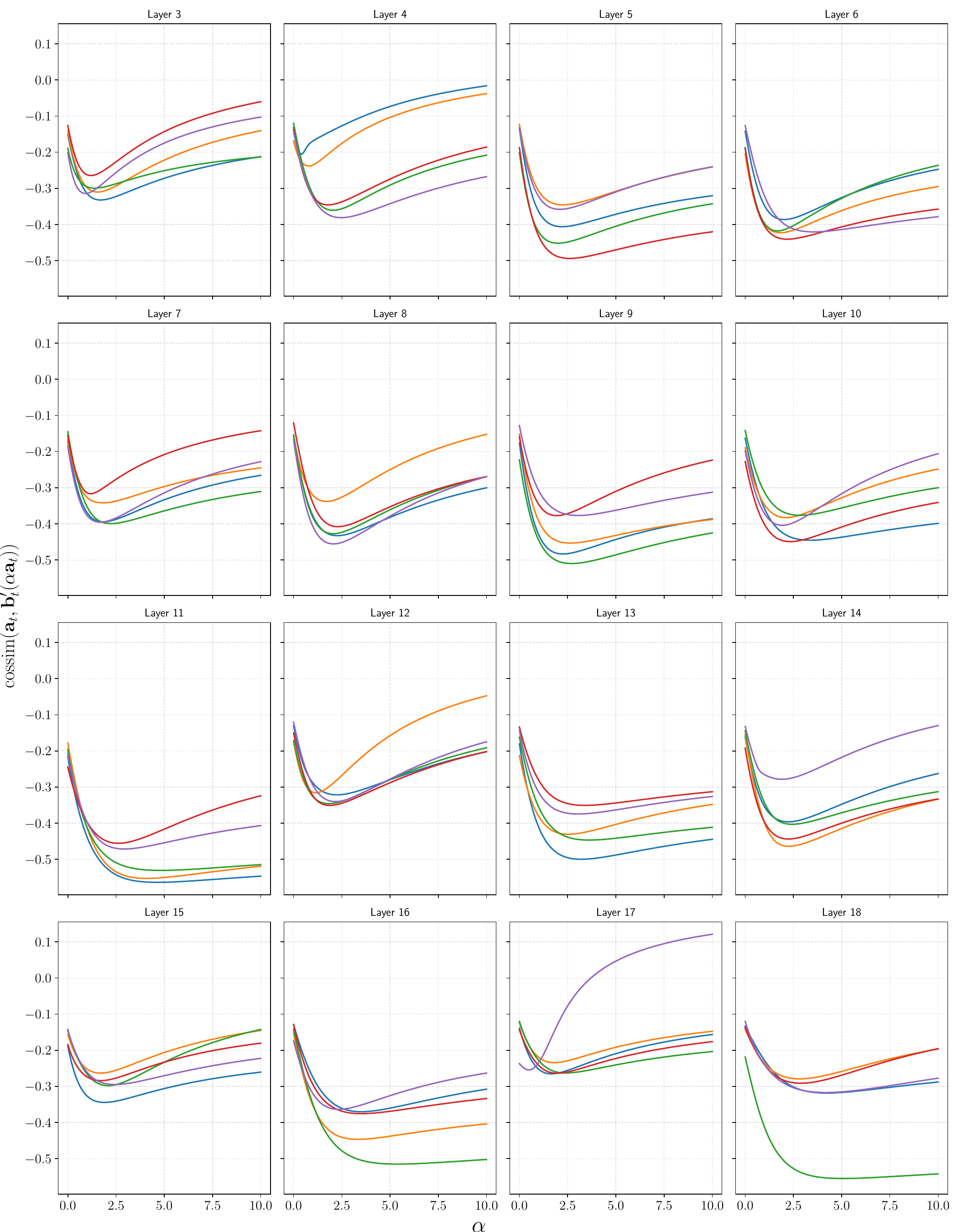}
    \caption{As we increase the previous layer more dramatically, we see TLCM's correction begin to diminish, aside from some outliers. For each layer, we sample 5 random TLCM curves and compute at $\alpha$ increments of $0.2$.}
    \label{fig:correction_diminishes}
\end{figure}

In Sec. \ref{sec:tlcm-adaptive}, we find that TLCM increases its correction as the prior layer is scaled. However, we find that as we scale the prior layer to extremely large values ($\alpha > 2$), the correction capacity of TLCM starts to diminish. One possible explanation is that at large enough perturbation magnitudes, the correction mechanism saturates. This observation is relevant to model steering interventions, in which a chosen feature vector is manually contributed to the residual stream and may need to exceed the correction capacity to take effect. See Figure \ref{fig:correction_diminishes} for plots of the correction beginning to diminish.

\pagebreak

\section{OLMo Training Checkpoint Experiment Details}
\begin{figure}
    \centering
    \includegraphics[trim={2.5mm 1mm 2mm 1mm},clip,width=\textwidth]{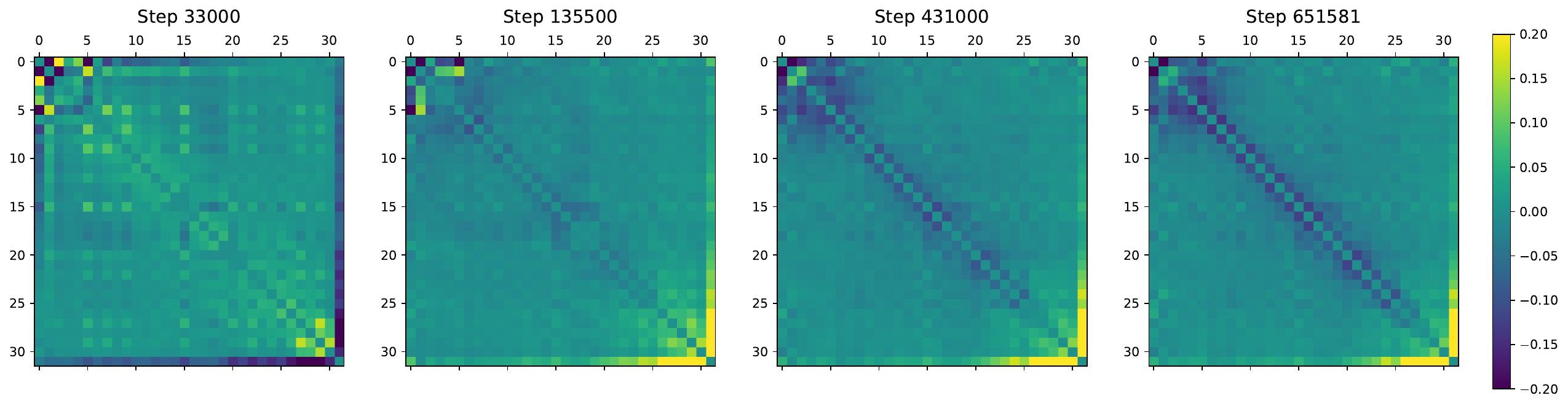}
    \caption{In OLMo 7B, TLCM emerges progressively during pretraining, with initial manifestation around step 135,500 (0.5T tokens). These four plots show $\mathbf{M}$ computed at different training checkpoints of OLMo 7B, with the rightmost plot representing the fully pretrained model. For visual clarity, we plot $\operatorname{clamp}(\mathbf{M}, -0.2, 0.2)$ and zero the diagonals.}
    \label{fig:olmo_checkpoints}
\end{figure}

On a corpus of about 500 tokens of PyTorch instructional content, we compute $\mathbf{M}$ on a handful of training checkpoints, shown in Figure \ref{fig:olmo_checkpoints}.

The corpus is below; beyond a high enough number of tokens, we find this to have little effect on the cosine similarity matrices and thus also the figures.

\begin{examplebox}
\begin{lstlisting}
**Using Convolutional Layers in PyTorch**
=====================================================

Convolutional layers are a fundamental component of convolutional neural networks (CNNs) used for image classification, object detection, and other computer vision tasks. In PyTorch, convolutional layers are implemented using the `nn.Conv2d` module.

**Creating a Convolutional Layer**
-------------------------------

To create a convolutional layer in PyTorch, you can use the following code:

```python
import torch
import torch.nn as nn

# Define the convolutional layer
conv_layer = nn.Conv2d(in_channels, out_channels, kernel_size, stride, padding)
```

*   `in_channels`: The number of input channels (e.g., 3 for RGB images).
*   `out_channels`: The number of output channels (e.g., 64 for a feature map).
*   `kernel_size`: The size of the convolutional kernel (e.g., 3x3).
*   `stride`: The stride of the convolutional kernel (e.g., 1).
*   `padding`: The amount of padding to apply (e.g., 1).

**Example Usage**
-----------------

Here's an example of using a convolutional layer in a PyTorch model:

```python
import torch
import torch.nn as nn

class ConvNet(nn.Module):
    def __init__(self):
        super(ConvNet, self).__init__()
        self.conv_layer = nn.Conv2d(3, 64, kernel_size=3, stride=1, padding=1)

    def forward(self, x):
        return torch.relu(self.conv_layer(x))

# Initialize the model and input tensor
model = ConvNet()
input_tensor = torch.randn(1, 3, 224, 224)

# Forward pass
output = model(input_tensor)
\end{lstlisting}
\end{examplebox}

\begin{figure}
    \centering
    \includegraphics[trim={2.5mm 1mm 2mm 1mm},clip,width=\textwidth]{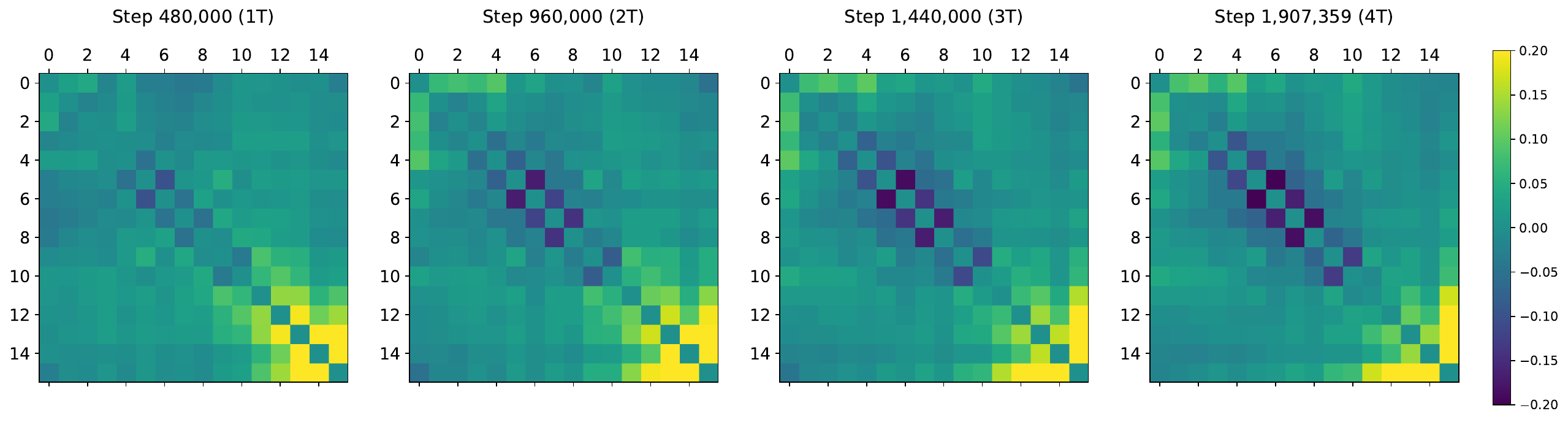}
    \caption{In OLMo 2 1B, TLCM also emerges progressively during pretraining. We plot figures at different step numbers and number of tokens (1, 2, 3, or 4 trillion tokens).}
    \label{fig:olmo2_1B_checkpoints}
\end{figure}

\begin{figure}
    \centering
    \includegraphics[trim={2.5mm 1mm 2mm 1mm},clip,width=\textwidth]{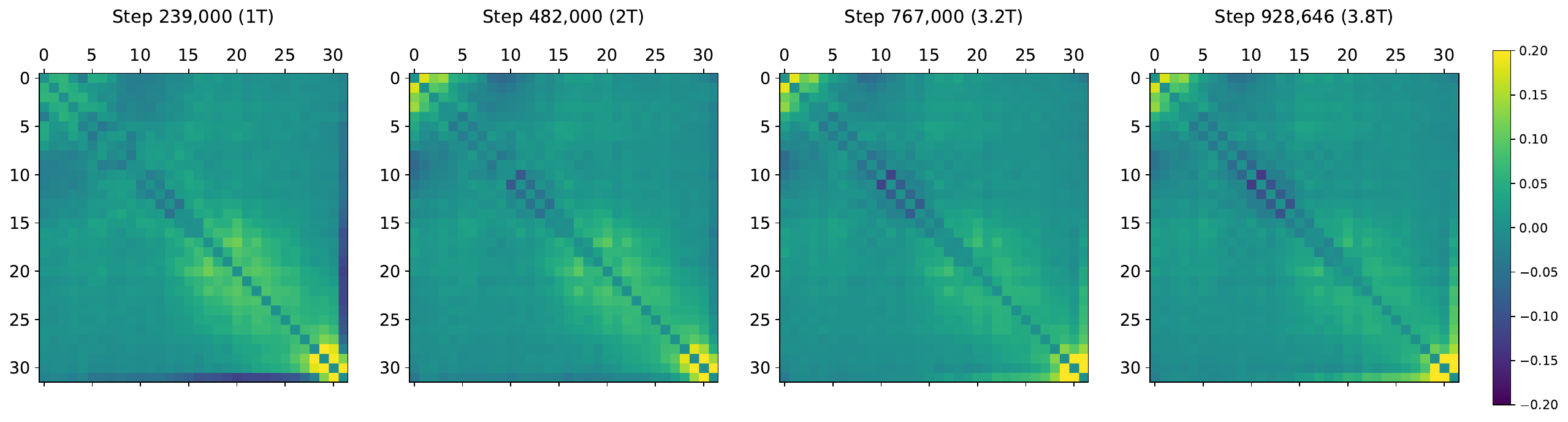}
    \caption{In OLMo 2 7B, TLCM also emerges progressively during pretraining. We plot figures at different step numbers and number of tokens.}
    \label{fig:olmo2_7B_checkpoints}
\end{figure}

\pagebreak
\section{Token-Level Correction Statistics}

\subsection{Experiment Prompts}

We use the following prompts to generate data for our experiment in Sec. \ref{sec:tlcm-tokens}:

\begin{examplebox}
\small
Write a blog post about the impact of remote work on urban real estate trends. \\
Write an essay on the psychological effects of social media on teenagers. \\
Write a report detailing the advancements in renewable energy technologies over the last decade. \\
Write an article about the rise of plant-based diets and their environmental benefits. \\
Write a memo to employees explaining the new company policy on cybersecurity measures. \\
Write a letter to a local council advocating for improved recycling facilities in the community. \\
Write a proposal for implementing a mindfulness program in elementary schools to enhance student well-being. \\
Write a blog post about the evolution of smart home technology and its implications for privacy. \\
Write an essay discussing the ethical considerations of genetic editing technologies. \\
Write a report on the economic impacts of the COVID-19 pandemic on small businesses. \\
Write an article about the significance of the James Webb Space Telescope's latest findings. \\
Write a memo outlining the steps for a successful digital transformation in a manufacturing company. \\
Write a letter to a senator expressing concerns about the proposed changes to healthcare laws. \\
Write a proposal for a community garden project to promote local food production and community engagement. \\
Write a blog post about the latest trends in artificial intelligence and machine learning. \\
Write an essay on the role of art therapy in mental health recovery. \\
Write a report assessing the potential of hydrogen fuel as an alternative energy source. \\
Write an article highlighting the importance of biodiversity conservation in combating climate change. \\
Write a memo to staff regarding the integration of a new project management software. \\
Write a letter to an editor expressing opinions on the local government's transportation plan. \\
Write a proposal for a telemedicine service to increase healthcare access in rural areas. \\
Write a blog post discussing the future of space tourism and its possible timeline. \\
Write an essay exploring the cultural significance of indigenous music. \\
Write a report on the trends in global unemployment rates and their implications for economic policy. \\
Write an article about the benefits and challenges of homeschooling. \\
Write a memo describing the company's strategy to address the upcoming industry regulations. \\
Write a letter to a non-profit organization offering to partner on an environmental initiative. \\
Write a proposal for an employee wellness program that includes both physical and mental health activities. \\
Write a blog post analyzing the impact of blockchain technology on financial services. \\
Write an essay on the historical impact of major pandemics on societal structures. \\
Write a report on the viability of vertical farming in urban environments. \\
Write an article about the challenges of maintaining data privacy in the age of IoT. \\
Write a memo to update company leadership on the progress of the quarterly goals. \\
Write a letter to a school board proposing the introduction of coding classes in middle schools. \\
Write a proposal for a local government initiative to support small businesses during economic downturns. \\
Write a blog post about the techniques and benefits of sustainable agriculture. \\
Write an essay on the influence of classical music on modern genres. \\
Write a report on consumer behavior changes in the automotive industry towards electric vehicles. \\
Write an article about the role of youth activism in shaping public policy.
\end{examplebox}

Continued on the next page.

\begin{examplebox}
\small
Write a memo detailing guidelines for handling customer data under new privacy laws. \\
Write a letter to the editor about the importance of public parks and open spaces. \\
Write a proposal for a new arts festival aiming to showcase local and international talent.\\
Write a blog post on the role of robotics in healthcare and potential ethical dilemmas. \\
Write an essay about the impact of climate change on marine ecosystems.\\
Write a report on strategies for managing workplace diversity in a global company. \\
Write an article on the resurgence of interest in vinyl records and analog music. \\
Write a memo to department heads about managing remote teams effectively.\\
Write a letter to a city planner regarding the need for improved pedestrian pathways.\\
Write a proposal for implementing a bike-sharing program in a mid-sized city. \\
Write a blog post about the future trends in education technology and their implications for learning. \\
Write a blog post about the growing popularity of mindfulness apps and their effectiveness. \\
Write an essay on the resurgence of traditional farming techniques in modern agriculture. \\
Write a report on the adoption of electric vehicles in major cities around the world. \\
Write an article about the psychological benefits of outdoor activities. \\
Write a memo to management detailing the steps to achieve carbon neutrality in the workplace by 2030. \\
Write a letter to a philanthropic organization requesting funding for a community tech hub. \\
Write a proposal for a series of workshops aimed at teaching digital literacy to seniors. \\
Write a blog post analyzing the impact of virtual reality on entertainment and media. \\
Write an essay discussing the philosophical implications of artificial intelligence surpassing human intelligence. \\
Write a report on the state of child nutrition programs in public schools. \\
Write an article about the role of drones in modern agriculture and their environmental impact. \\
Write a memo regarding the implementation of a flexible work schedule to enhance employee productivity. \\
Write a letter to a government official advocating for stricter air pollution regulations. \\
Write a proposal for a new public library with advanced digital resources. \\
Write a blog post about the importance of cybersecurity in the age of cloud computing. \\
Write an essay exploring the historical role of spices in global trade. \\
Write a report on the effectiveness of recent public health campaigns on smoking cessation. \\
Write an article on the growing trend of micro-living and tiny homes. \\
Write a memo introducing a new internal team dedicated to innovation and strategic initiatives. \\
Write a letter to parents outlining the new curriculum changes in a local school district. \\
Write a proposal for a mobile health clinic to serve underserved areas. \\
Write a blog post about the use of big data in personalized medicine. \\
Write an essay on the evolution of language in the digital age. \\
Write a report detailing the economic impact of cultural festivals on local communities. \\
Write an article on the significance of urban green spaces for mental health. \\
Write a memo to staff about upcoming training opportunities in advanced analytics. \\
Write a letter to the editor discussing the need for more inclusive sports programs in schools. \\
Write a proposal for an annual technology conference focusing on sustainability innovations. \\
Write a blog post about the effects of music therapy on Alzheimer's patients. \\
Write an essay examining the influence of video games on cognitive development. \\
Write a report on the future of nuclear energy and its role in combating climate change. \\
Write an article about the revival of handcrafts and their market in the modern economy. \\
Write a memo outlining the benefits of adopting a four-day workweek. \\
Write a letter to a university proposing a partnership for a community-based research project. \\
Write a proposal for developing a pedestrian-friendly zone in the downtown area. \\
Write a blog post on innovative approaches to waste management in urban settings. \\
Write an essay about the socio-economic impacts of migration on urban development. \\
Write a report on the adoption and regulation of cryptocurrencies in different countries. \\
Write an article on how to prepare pets for the arrival of a new baby. \\
Write a memo discussing the integration of virtual assistants into customer service. \\
Write a letter to a historical society proposing a project to digitize and preserve ancient manuscripts. \\
Write a proposal for a fitness program aimed at improving the health of office workers. \\
Write a blog post about the role of augmented reality in modern education. \\
Write an essay on the impact of global trade policies on developing economies. \\
Write a report analyzing the trends in youth sports and their benefits to communities. \\
Write an article about the ethical considerations in wildlife photography. \\
Write a memo to update the company on the progress of the diversity and inclusion initiative. \\
Write a letter to an NGO outlining a proposal for a joint clean water project in rural areas. \\
Write a proposal for a digital art exhibition featuring interactive installations. \\
Write a blog post discussing the future of autonomous public transit systems and their societal impacts.
\end{examplebox}

\subsection{Correction Counts}

Each tuple follows the format: (token, average TLCM activation count). We remove a few hundred tokens from the middle of this distribution due to space constraints. We assess pairwise differences in mean activation counts using two-sided Welch's $t$-tests on per-occurrence TLCM activation counts, allowing unequal variances across tokens.

\begin{table}[h!]
\centering
\caption{Llama 3.1 8B Instruct: Top 50 tokens with the highest average number of TLCM activations. For each token, we list the average number of times a TLCM activations occurs on the given token, aggregated across 100 long documents. Finally, we list the standard error of the mean and the number of occurrences of the token across the corpus.}
\label{table:token_correction_counts_llama_top}
\begin{tabular}{llll}
\toprule
\textbf{Token} & \textbf{Mean Activations} & \textbf{Std err. of mean} & \textbf{\# occurrences} \\
\midrule
\texttt{202} & 16.66 & 0.16 & 265 \\
\texttt{ Jul} & 14.00 & 0.00 & 100 \\
\texttt{26} & 13.90 & 0.04 & 112 \\
\texttt{]$\backslash$n} & 13.15 & 0.12 & 131 \\
\texttt{Today} & 13.00 & 0.00 & 100 \\
\texttt{$\backslash$n$\backslash$n} & 12.88 & 0.11 & 485 \\
\texttt{,$\backslash$n$\backslash$n} & 12.85 & 0.16 & 26 \\
\texttt{ $\$$} & 12.62 & 0.19 & 55 \\
\texttt{<space>} & 12.58 & 0.07 & 1276 \\
\texttt{ Name} & 12.54 & 0.10 & 100 \\
\texttt{ at} & 12.44 & 0.24 & 41 \\
\texttt{$\backslash$n} & 12.41 & 0.10 & 228 \\
\texttt{[} & 12.23 & 0.08 & 164 \\
\texttt{assistant} & 12.20 & 0.08 & 100 \\
\texttt{ State} & 12.15 & 0.32 & 26 \\
\texttt{]} & 12.08 & 0.14 & 49 \\
\texttt{$\backslash$t} & 12.07 & 0.20 & 68 \\
\texttt{Date} & 12.03 & 0.24 & 30 \\
\texttt{ Address} & 12.03 & 0.26 & 35 \\
\texttt{user} & 12.00 & 0.00 & 100 \\
\texttt{]$\backslash$n$\backslash$n} & 12.00 & 0.19 & 39 \\
\texttt{ Date} & 11.99 & 0.14 & 204 \\
\texttt{4} & 11.92 & 0.12 & 338 \\
\texttt{Your} & 11.81 & 0.12 & 103 \\
\texttt{ over} & 11.74 & 0.26 & 35 \\
\texttt{<space><space><space>} & 11.68 & 0.19 & 57 \\
\texttt{-} & 11.68 & 0.13 & 112 \\
\texttt{$\%$} & 11.65 & 0.14 & 52 \\
\texttt{:} & 11.58 & 0.08 & 606 \\
\texttt{ D} & 11.46 & 0.23 & 28 \\
\texttt{ high} & 11.46 & 0.23 & 39 \\
\texttt{ from} & 11.44 & 0.14 & 109 \\
\texttt{ make} & 11.44 & 0.20 & 50 \\
\texttt{ you} & 11.43 & 0.15 & 96 \\
\texttt{ access} & 11.40 & 0.15 & 89 \\
\texttt{ take} & 11.37 & 0.21 & 30 \\
\texttt{ between} & 11.30 & 0.26 & 27 \\
\texttt{ [} & 11.28 & 0.12 & 108 \\
\texttt{City} & 11.24 & 0.30 & 38 \\
\texttt{ not} & 11.23 & 0.17 & 52 \\
\texttt{ well} & 11.23 & 0.15 & 70 \\
\texttt{ need} & 11.22 & 0.19 & 55 \\
\texttt{Thank} & 11.19 & 0.18 & 27 \\
\texttt{1} & 11.18 & 0.07 & 295 \\
\texttt{ your} & 11.10 & 0.13 & 100 \\
\texttt{ such} & 11.05 & 0.10 & 157 \\
\texttt{ up} & 11.03 & 0.27 & 39 \\
\texttt{ Knowledge} & 11.00 & 0.00 & 100 \\
\texttt{Write} & 11.00 & 0.00 & 100 \\
\texttt{ long} & 11.00 & 0.27 & 26 \\
\bottomrule
\end{tabular}
\end{table}

\begin{table}[h!]
\centering
\caption{Llama 3.1 8B Instruct: Top 50 tokens with the lowest average number of TLCM activations. For each token, we list the average number of times a TLCM activations occurs on the given token, aggregated across 100 long documents. Finally, we list the standard error of the mean and the number of occurrences of the token across the corpus.}
\label{table:token_correction_counts_llama_bottom}
\begin{tabular}{llll}
\toprule
\textbf{Token} & \textbf{Mean Activations} & \textbf{Std err. of mean} & \textbf{\# occurrences} \\
\midrule
\texttt{ program} & 8.52 & 0.13 & 97 \\
\texttt{ coding} & 8.50 & 0.24 & 26 \\
\texttt{ report} & 8.49 & 0.28 & 35 \\
\texttt{ guidelines} & 8.44 & 0.26 & 27 \\
\texttt{ AI} & 8.42 & 0.19 & 57 \\
\texttt{ home} & 8.41 & 0.27 & 34 \\
\texttt{ efficiency} & 8.41 & 0.24 & 37 \\
\texttt{ art} & 8.38 & 0.30 & 32 \\
\texttt{ This} & 8.37 & 0.07 & 155 \\
\texttt{ organizations} & 8.33 & 0.22 & 36 \\
\texttt{ community} & 8.33 & 0.09 & 160 \\
\texttt{ization} & 8.23 & 0.27 & 39 \\
\texttt{**:} & 8.22 & 0.06 & 601 \\
\texttt{ media} & 8.22 & 0.22 & 60 \\
\texttt{ regulations} & 8.22 & 0.19 & 51 \\
\texttt{-being} & 8.20 & 0.20 & 59 \\
\texttt{ engagement} & 8.17 & 0.16 & 63 \\
\texttt{ environmental} & 8.17 & 0.20 & 42 \\
\texttt{-based} & 8.15 & 0.23 & 40 \\
\texttt{ sustainable} & 8.15 & 0.11 & 89 \\
\texttt{ urban} & 8.14 & 0.14 & 70 \\
\texttt{ learning} & 8.12 & 0.18 & 80 \\
\texttt{ infrastructure} & 8.12 & 0.20 & 49 \\
\texttt{ IoT} & 8.10 & 0.25 & 31 \\
\texttt{ should} & 8.08 & 0.14 & 59 \\
\texttt{ interactive} & 8.08 & 0.25 & 26 \\
\texttt{ diversity} & 8.07 & 0.24 & 28 \\
\texttt{ classical} & 8.07 & 0.24 & 30 \\
\texttt{ challenges} & 8.07 & 0.13 & 92 \\
\texttt{ cities} & 8.03 & 0.18 & 40 \\
\texttt{ mindfulness} & 8.00 & 0.29 & 33 \\
\texttt{ indigenous} & 8.00 & 0.17 & 27 \\
\texttt{ agriculture} & 7.96 & 0.21 & 50 \\
\texttt{ VR} & 7.93 & 0.26 & 29 \\
\texttt{ workshops} & 7.93 & 0.25 & 28 \\
\texttt{ By} & 7.92 & 0.11 & 77 \\
\texttt{ therapy} & 7.89 & 0.24 & 37 \\
\texttt{ trends} & 7.89 & 0.27 & 27 \\
\texttt{ innovative} & 7.89 & 0.25 & 27 \\
\texttt{ innovation} & 7.89 & 0.18 & 35 \\
\texttt{ proposal} & 7.86 & 0.42 & 37 \\
\texttt{ cognitive} & 7.85 & 0.22 & 26 \\
\texttt{ inclusive} & 7.66 & 0.17 & 41 \\
\texttt{ sustainability} & 7.63 & 0.23 & 35 \\
\texttt{ tourism} & 7.52 & 0.23 & 27 \\
\texttt{ Cities} & 7.33 & 0.22 & 27 \\
\texttt{ting} & 7.03 & 0.05 & 102 \\
\texttt{ blog} & 6.50 & 0.55 & 28 \\
\texttt{system} & 3.00 & 0.00 & 100 \\
\texttt{$\texttt{<|begin\_of\_text|>}$} & 0.00 & 0.00 & 100 \\
\bottomrule
\end{tabular}
\end{table}

\begin{table}[h!]
\centering
\caption{Gemma 2 2B Instruct: Top 50 tokens with the highest average number of TLCM activations. For each token, we list the average number of times a TLCM activations occurs on the given token, aggregated across 100 long documents. Finally, we list the standard error of the mean and the number of occurrences of the token across the corpus.}
\label{table:token_correction_counts_gemma_top}
\begin{tabular}{llll}
\toprule
\textbf{Token} & \textbf{Mean Activations} & \textbf{Std err. of mean} & \textbf{\# occurrences} \\
\midrule
\texttt{<space>} & 17.89 & 0.06 & 623 \\
\texttt{4} & 17.31 & 0.11 & 366 \\
\texttt{6} & 17.09 & 0.13 & 160 \\
\texttt{$\backslash$t} & 16.56 & 0.26 & 68 \\
\texttt{2} & 16.56 & 0.06 & 1066 \\
\texttt{'} & 16.48 & 0.14 & 242 \\
\texttt{]} & 16.39 & 0.13 & 218 \\
\texttt{<space><space>} & 16.36 & 0.09 & 657 \\
\texttt{7} & 16.17 & 0.32 & 35 \\
\texttt{<space><space><space><space>} & 16.00 & 0.27 & 57 \\
\texttt{$\%$} & 15.63 & 0.29 & 57 \\
\texttt{3} & 15.61 & 0.08 & 424 \\
\texttt{$\backslash$n} & 15.58 & 0.05 & 1289 \\
\texttt{5} & 15.43 & 0.12 & 183 \\
\texttt{-} & 15.42 & 0.08 & 840 \\
\texttt{0} & 15.32 & 0.06 & 627 \\
\texttt{],} & 15.27 & 0.41 & 33 \\
\texttt{Thank} & 15.22 & 0.13 & 27 \\
\texttt{$\backslash$n$\backslash$n} & 15.16 & 0.06 & 1915 \\
\texttt{ $\$$} & 15.09 & 0.25 & 55 \\
\texttt{ such} & 15.06 & 0.13 & 157 \\
\texttt{1} & 15.02 & 0.13 & 434 \\
\texttt{Today} & 15.00 & 0.00 & 100 \\
\texttt{9} & 14.96 & 0.44 & 50 \\
\texttt{.} & 14.87 & 0.04 & 3713 \\
\texttt{ Address} & 14.74 & 0.17 & 34 \\
\texttt{ recent} & 14.61 & 0.29 & 33 \\
\texttt{"} & 14.59 & 0.30 & 49 \\
\texttt{).} & 14.57 & 0.28 & 49 \\
\texttt{ Date} & 14.49 & 0.04 & 204 \\
\texttt{:} & 14.49 & 0.10 & 476 \\
\texttt{ long} & 14.46 & 0.37 & 26 \\
\texttt{)} & 14.37 & 0.22 & 84 \\
\texttt{ them} & 14.27 & 0.35 & 60 \\
\texttt{Date} & 14.17 & 0.18 & 30 \\
\texttt{[} & 14.14 & 0.14 & 166 \\
\texttt{ members} & 14.10 & 0.43 & 30 \\
\texttt{ sense} & 13.97 & 0.30 & 30 \\
\texttt{ (} & 13.90 & 0.17 & 195 \\
\texttt{ modern} & 13.81 & 0.32 & 32 \\
\texttt{ "} & 13.79 & 0.23 & 43 \\
\texttt{City} & 13.76 & 0.30 & 38 \\
\texttt{ over} & 13.71 & 0.27 & 34 \\
\texttt{ years} & 13.70 & 0.35 & 40 \\
\texttt{ at} & 13.68 & 0.35 & 41 \\
\texttt{ Name} & 13.66 & 0.14 & 100 \\
\texttt{ countries} & 13.59 & 0.36 & 37 \\
\texttt{assistant} & 13.55 & 0.14 & 100 \\
\texttt{ world} & 13.52 & 0.28 & 54 \\
\texttt{ led} & 13.50 & 0.28 & 44 \\
\bottomrule
\end{tabular}
\end{table}

\begin{table}[h!]
\centering
\caption{Gemma 2 2B Instruct: Top 50 tokens with the lowest average number of TLCM activations. For each token, we list the average number of times a TLCM activations occurs on the given token, aggregated across 100 long documents. Finally, we list the standard error of the mean and the number of occurrences of the token across the corpus.}
\label{table:token_correction_counts_gemma_bottom}
\begin{tabular}{llll}
\toprule
\textbf{Token} & \textbf{Mean Activations} & \textbf{Std err. of mean} & \textbf{\# occurrences} \\
\midrule
\texttt{ sustainable} & 10.25 & 0.18 & 89 \\
\texttt{ spaces} & 10.24 & 0.29 & 46 \\
\texttt{ plan} & 10.23 & 0.39 & 39 \\
\texttt{ EV} & 10.23 & 0.26 & 26 \\
\texttt{ several} & 10.22 & 0.31 & 27 \\
\texttt{ understanding} & 10.22 & 0.33 & 32 \\
\texttt{ growing} & 10.21 & 0.24 & 38 \\
\texttt{ online} & 10.21 & 0.17 & 43 \\
\texttt{ reduced} & 10.19 & 0.28 & 26 \\
\texttt{ learning} & 10.19 & 0.20 & 80 \\
\texttt{ environmental} & 10.14 & 0.18 & 42 \\
\texttt{ training} & 10.14 & 0.27 & 59 \\
\texttt{ create} & 10.10 & 0.22 & 69 \\
\texttt{ implementing} & 10.10 & 0.35 & 30 \\
\texttt{ mental} & 10.09 & 0.21 & 53 \\
\texttt{ AI} & 10.09 & 0.21 & 57 \\
\texttt{ indigenous} & 10.07 & 0.29 & 27 \\
\texttt{ local} & 10.07 & 0.16 & 110 \\
\texttt{ significant} & 10.06 & 0.16 & 143 \\
\texttt{ comprehensive} & 10.05 & 0.22 & 39 \\
\texttt{ together} & 10.04 & 0.41 & 26 \\
\texttt{ blog} & 10.00 & 0.07 & 28 \\
\texttt{ address} & 10.00 & 0.33 & 41 \\
\texttt{ innovative} & 9.96 & 0.29 & 27 \\
\texttt{ promoting} & 9.94 & 0.24 & 47 \\
\texttt{ post} & 9.94 & 0.22 & 31 \\
\texttt{ develop} & 9.89 & 0.25 & 35 \\
\texttt{ benefits} & 9.87 & 0.19 & 105 \\
\texttt{ VR} & 9.86 & 0.28 & 29 \\
\texttt{ community} & 9.83 & 0.17 & 160 \\
\texttt{ awareness} & 9.80 & 0.41 & 30 \\
\texttt{ prioritize} & 9.79 & 0.29 & 28 \\
\texttt{ promote} & 9.74 & 0.15 & 102 \\
\texttt{ progress} & 9.72 & 0.38 & 29 \\
\texttt{ following} & 9.70 & 0.31 & 27 \\
\texttt{ clear} & 9.69 & 0.33 & 26 \\
\texttt{ concerns} & 9.67 & 0.32 & 48 \\
\texttt{ approach} & 9.65 & 0.34 & 34 \\
\texttt{ complex} & 9.61 & 0.27 & 38 \\
\texttt{ By} & 9.58 & 0.17 & 77 \\
\texttt{ challenges} & 9.53 & 0.25 & 92 \\
\texttt{ improved} & 9.52 & 0.28 & 33 \\
\texttt{ interactive} & 9.42 & 0.38 & 26 \\
\texttt{ enhance} & 8.97 & 0.34 & 35 \\
\texttt{ improve} & 8.91 & 0.20 & 70 \\
\texttt{ explore} & 8.74 & 0.29 & 39 \\
\texttt{ feedback} & 8.70 & 0.28 & 27 \\
\texttt{ mitigate} & 7.88 & 0.37 & 26 \\
\texttt{Write} & 7.00 & 0.00 & 100 \\
\texttt{<bos>} & 1.00 & 0.00 & 100 \\
\bottomrule
\end{tabular}
\end{table}

%% file: sections/5_layernorm_blindness.tex
One natural cause of TLCM is RMSNorm. RMSNorm normalizes the input to the attention and MLP sublayers in nearly all of the open-source models analyzed.  
Formally, for an input $\mathbf{x} \in \mathbb{R}^{d_\text{m}}$ to an attention or MLP sublayers and a learnable parameter vector $\mathbf{g}$, RMSNorm is defined as:
$$\overline{\mathbf{x}} = \frac{\mathbf{x}}{\operatorname{RMS}(\mathbf{x})} \odot \mathbf{g} \qquad \qquad \operatorname{RMS}(\mathbf{x}) = \frac{1}{\sqrt{d_\text{m}}} \|\mathbf{x}\|_2,$$
where $d_\text{m}$ is the dimension of the residual stream. 

Crucially, the use of RMSNorm implies that both the attention and MLP sublayers have \emph{layernorm blindness}; they are blind to the norm of the residual stream. This blindness is significant because these sublayers predict contributions to the residual stream (i.e. $\mathbf{x}_i + \operatorname{Sublayer}(\mathbf{x}_i)$), whose relative impacts depend on the the residual streams current magnitude. 
Without visibility into the residual stream norm, attention and MLP sublayers risk under-contributing when the norm is high, which potentially leads to their contributions being overshadowed. This could encourage over-contribution behaviors followed by correction, which is consistent with TLCM. 

However, we find that \emph{LayerNorm blindness alone does not fully explain the  TLCM}. This is because as discussed in Sec. \ref{sec:jacobian-selective}, the correction mechanism does not entirely reverse $\mathbf{a}_t$, contrary to what would be expected if RMSNorm were the primary cause. But perhaps more critically, models trained with alternative RMSNorm implementations continue to exhibit the mechanism:

\textbf{Gemma.} The sublayers in Gemma 1 were trained using pre-LayerNorm as described above. In contrast, Gemma 2 utilized a hybrid approach, employing both pre-LayerNorm and post-LayerNorm:
$$\mathbf{x}_{i+1} = \mathbf{x}_i + \operatorname{LayerNorm}(\operatorname{Sublayer}(\operatorname{LayerNorm}(\mathbf{x}_i))).$$
The addition of post-LayerNorm should, in principle, make the residual stream norm more predictable. However, empirical results show that the correction mechanism remains robust in Gemma 2. Refer to Figure \ref{fig:rmsnorm} for a comparison.

\textbf{OLMo.} The original OLMo models employed pre-LayerNorm exclusively. In the OLMo 2 series, pre-LayerNorm was replaced with post-LayerNorm and QK norm. Despite this architectural change, the correction mechanism persists strongly in OLMo 2, as shown in Figure \ref{fig:rmsnorm}.
